\documentclass[a4,notitlepage,12pt]{jedm}

\usepackage[table]{xcolor}
\usepackage{url}
\usepackage{graphicx}
\usepackage{fancyhdr}
\usepackage{multirow}
\usepackage{eurosym}
\usepackage{graphicx}

\usepackage{algorithm}
\usepackage{algorithmic}

\usepackage{color}

\def\artvolume{7}
\def\artno{2}
\def\artmonthyear{2015}
\def\startpage{20}               

\begin{document}

\date{}
\title{Multi-Armed Bandits for\\ Intelligent Tutoring Systems}



\author{{\large Benjamin Clement}, \and {\large Didier Roy},  \and {\large Pierre-Yves Oudeyer}, \and {\large Manuel Lopes}\\manuel.lopes@inria.fr\\\url{http://flowers.inria.fr}\\Inria, Bordeaux, France}

\maketitle

\thispagestyle{fancy}

\begin{abstract}
We present an approach to Intelligent Tutoring Systems which adaptively personalizes sequences of learning activities to maximize skills acquired by students, taking into account the limited time and motivational resources. At a given point in time, the system proposes to the students the activity which makes them progress faster. We introduce two algorithms that rely on the empirical estimation of the learning progress, \textbf{RiARiT} that uses information about the difficulty of each exercise and \textbf{ZPDES} that uses much less knowledge about the problem.

The system is based on the combination of three approaches. First, it leverages recent models of intrinsically motivated learning by transposing them to active teaching, relying on empirical estimation of learning progress provided by specific activities to particular students. Second, it uses state-of-the-art Multi-Arm Bandit (MAB) techniques to efficiently manage the exploration/exploitation challenge of this optimization process. Third, it leverages expert knowledge to constrain and bootstrap initial exploration of the MAB, while requiring only coarse guidance information of the expert and allowing the system to deal with didactic gaps in its knowledge. The system is evaluated in a scenario where 7-8 year old schoolchildren learn how to decompose numbers while manipulating money. Systematic experiments are presented with simulated students, followed by results of a user study across a population of 400 school children.
\end{abstract}


\section{Introduction}
\label{sec:Introduction}

Intelligent Tutoring Systems (ITS) have been proposed to make education more accessible, more effective, and as a way to provide useful objective metrics on learning \cite{anderson1995cognitive,koedinger1997intelligent,nkambou2010advances}.

In general an ITS requires a cognitive and a student model, but in this work we will focus on the \textit{tutoring model}, that is how to choose the activities that provide a better learning experience based on the estimation of the student competence levels and progression, and little knowledge about the cognitive and student models. We can imagine a student wanting to acquire many different skills, e.g. adding, subtracting and multiplying numbers. A teacher can help students by proposing activities such as: multiple choice questions, abstract operations to compute with a pencil, games where items need to be counted through manipulation, videos, or others. The challenge is to decide what is the optimal sequence of activities that maximizes the average competence level over all skills. 

This is a difficult question for a teacher for at least three reasons. First, time resources are typically limited, where both studentds and teachers have a limited budget of time to allocate for practicing activities. Second, motivational resources are also limited, especially for the student, who will learn efficiently only if he is psychologically engaged in the activities.
Third, because of the individual differences between students, a sequence that is optimal for one may be inefficient for another student. 

Our main design principles, when compared to other ITS systems, are the following:

\paragraph{Weaker dependency on the cognitive and student model}
Given students' particularities, it is often highly difficult or impossible for a teacher to understand all the difficulties and strengths of individual students and thus predict which activities provide them with maximal learning progress.  Even when using automatic methods there are several challenges in identifying parameters that best describe each individual student \cite{beck2007identifiability,beck13limits,lee12teachindividual}. Because of this, we consider that it is important to be as independent as possible of a pre-defined population wide cognitive and student model. Instead we must adapt and estimate online the characteristics of each individual student \cite{Clement2014edm,Lopes12ssp}.

\paragraph{Efficient Optimization Methods}
We want methods that do not make specific assumptions about how students learn and only require information about the estimated learning progress of each activity. For this, we will rely on efficient online methods, multi-armed bandits, that are able to explore different activities to estimate the progress that they can give to each particular student, and then they exploit the ones that are best to improve students' learning. We present these in opposition to other methods that consider offline optimization considering population wide, and not individualized, parameters.

\paragraph{More Motivating Experience}
Our approach considers that exercises which are currently providing higher learning progress must be the ones proposed. This allows not only to use more efficient optimization algorithms but also to provide a more motivating experience to students. Several strands of work in psychology \cite{berlyne1960conflict} and neuroscience \cite{gottlieb2013information} have argued that the human brain feels intrinsic pleasure in practicing activities of optimal difficulty or challenge, i.e. neither too easy nor too difficult, but slightly beyond the current abilities. This type of activities have been described as the zone of proximal development where children can improve with small guidance \cite{lee2005signifying,Luckin2001k5} and the concept of flow where people feel more engaged in activity slightly higher than their current level \cite{csikszentmihalyi1992optimal}. This follows well known instructional design methodologies \cite{gagne1974principles,Luckin2001k5} and concords with theories of intrinsic motivation which clearly suggest motivation and learning improve if exercises are proposed at levels that are only slightly higher than the current level \cite{habgood2011motivating,engeser2008flow}.

If there are many activities we will need to explore all of them in order to estimate their impact on each knowledge component. Such exploration will be very time consuming and will provide under-performing learning sequences. Instead we allow our algorithms to be initialized with a canonical learning sequence upon which the algorithms can optimize. We provide the teachers a simple means to specify an initial learning sequence to make this process simpler.

~\\
Our main contribution is the use of multi-armed bandit algorithms for ITS. These algorithms allow a true personalized learning experience relying on little domain knowledge. Depending on the proposed algorithm, this knowledge includes only coarse pedagogical constraints and potentially a relation between activities and knowledge components. Our results show that this approach achieves comparable, and in some cases better, learning results than the sequence created by an expert teacher. This paper extends some of the results already published \cite{Clement2014edm} by providing much more details and by including complete user studies. 

In the following section, we review the related work, and present the methodological and algorithmic details of the proposed algorithms. Our approaches assume that an instructional expert defines a set of skills to acquire, a set of potential activities/exercises to practice and, if necessary, coarse constraints on the pedagogical sequence. Our first approach uses very little knowledge about the problem and is inspired by the zone of proximal development and the empirical estimation of learning progress hence the name ``Zone of Proximal Development and Empirical Success'' (\textbf{ZPDES}). Our second approach further assumes the existence of a simple relation between the activities and the skills. Then, at any given point in time, the system estimates the learning progress obtained for each activity by the student. The system then proposes to the student the activities which provide an higher learning progress, hence the name of the algorithm: the ``Right Activity at the Right Time'' (\textbf{RiARiT}). 

Finally, we present two experiments to evaluate our algorithms. In a first experiment, we conduct systematic statistical studies of the impact of our approaches over a population of simulated students. Then we present a real-world experiment where the approach is implemented as a tablet application used for learning number decomposition while using money. The experiment involves 400 children (7-8 year old) from 11 schools.  The effectiveness of our algorithms is measured by the comparison of their output to a teaching sequence handcrafted and validated by an expert.

\section{Related Work - Optimizing Teaching Sequences using Machine Learning}

There have been several approaches to optimize teaching sequences. Some approaches are based on hand-made optimization and on pedagogical theory, experience and domain knowledge. There are many works that followed this line but the approaches more relevant to the work presented in this article are those where the optimization is made automatically without particular assumptions about the students or the knowledge domain. This is a very active line of work, and approaches vary in their assumptions about the knowledge domain, goals in terms of personalization and availability of students' data. See \citeN{koedinger1997intelligent}, \citeN{koedinger13ITSsuvery}, and \citeN{nkambou2010advances} for a discussion on such topics.

The framework of partial-observable Markov decision process (POMDP) has been proposed to select the optimal activities to propose to the students based on the estimation of their level of acquisition of each skill \cite{rafferty2011faster}. In general the solution to a POMDP is a difficult problem and approximate solutions have been proposed using the concept of envelope states \cite{brunskill2012rapid} that, instead of tracking the full knowledge units, considers groups of units. In most cases the tutoring model incorporates the student model inside. For instance, in approaches based on POMDPs, the optimization of teaching sequences is made by assuming that all students learn in the same way.

These approaches are potentially optimal but they require good student and cognitive models. POMDP will plan the optimal trajectory based on that model of the students. For this model, many approaches rely on Knowledge Tracing methods \cite{corbett1994knowledge}, or variants, and some methods already try to estimate those parameters form data \cite{gonzalez2012dynamic,Baker2008more,gonzalez2014general,dhanani2014parameters}. Typically, these models have many parameters, and identifying all such parameters for a single student is a very hard problem due to the lack of data, the intractability of the problem, and the lack of identifiability of many parameters \cite{beck2007identifiability,beck13limits}. This often results in models which are inaccurate in practice. Another problem is that these planning methods are for a population of students and not for a particular student and this has already been proven to be suboptimal \cite{lee12teachindividual}.

Other approaches used reinforcement learning to provide hints during problem solving \cite{barnes2011using}, and to improve the adaptation of pedagogical strategies \cite{chi2011empirically} or used bayesian networks to model and decide how to help students \cite{gertner1998procedural}. Other approaches consider a global optimization of the pedagogical sequence based on data from all the student using ant colony optimization algorithms \cite{semet2003artificial}, but can not provide a personalized sequence.

Several authors already considered the design of ITS based on the use of the zone-of-proximal-development based on educational design principles \cite{Luckin2001k5} or based on data mining approaches \cite{schatten2014matrix}. Our work differs from these approaches in that the ZPD is defined approximately by an expert and then the optimization algorithms will adjust this zone based on the answers and learning progress of the students.


%

\section{Teaching Scenario}
\label{sec:TeachingScenario}

In this section, we present the teaching scenario we use and the experimental protocol followed in the user studies. This scenario is about learning how to decompose numbers while using money, typically targeted to 7-8 years old students. Such a scenario was chosen due to its simplicity but having enough richness to enable different learning/teaching trajectories to impact particular students differentially. Furthermore, combining number and money manipulation is a way to instantiate abstract knowledge into a practical useful real-world scenario.

\begin{figure}[tbp]
  \centering
    \includegraphics[width=0.45\columnwidth]{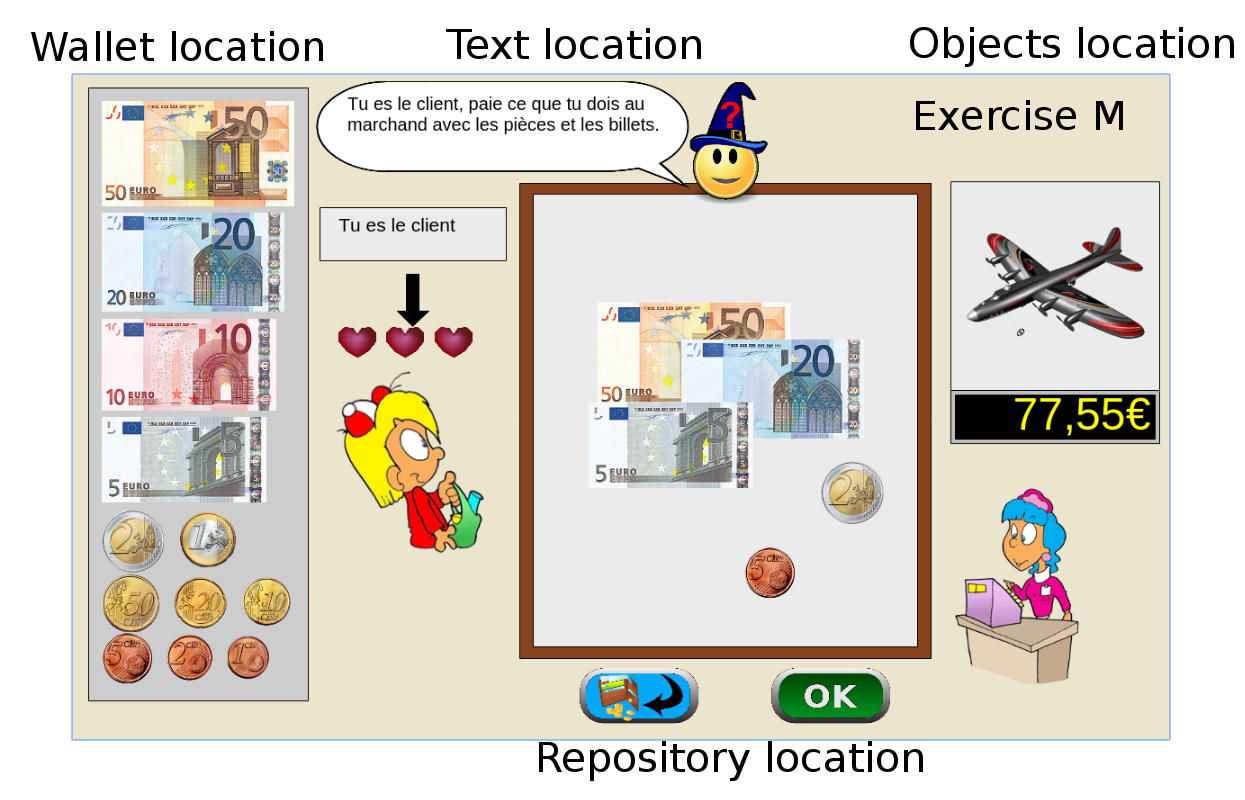}
    \includegraphics[width=0.45\columnwidth]{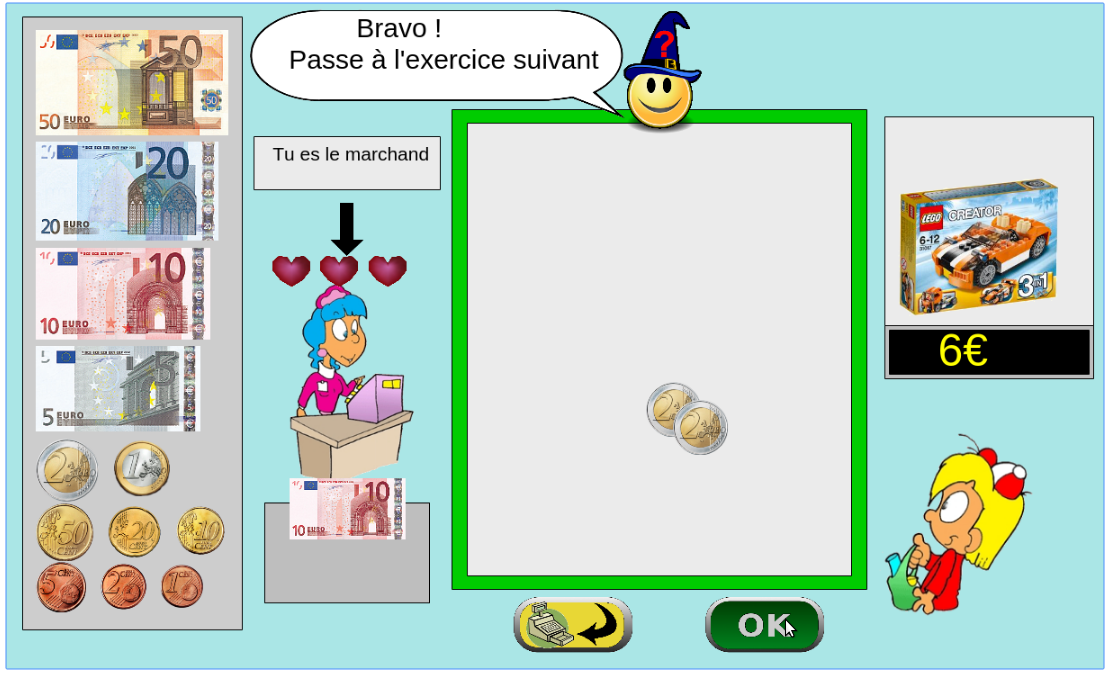}
    \includegraphics[width=0.45\columnwidth]{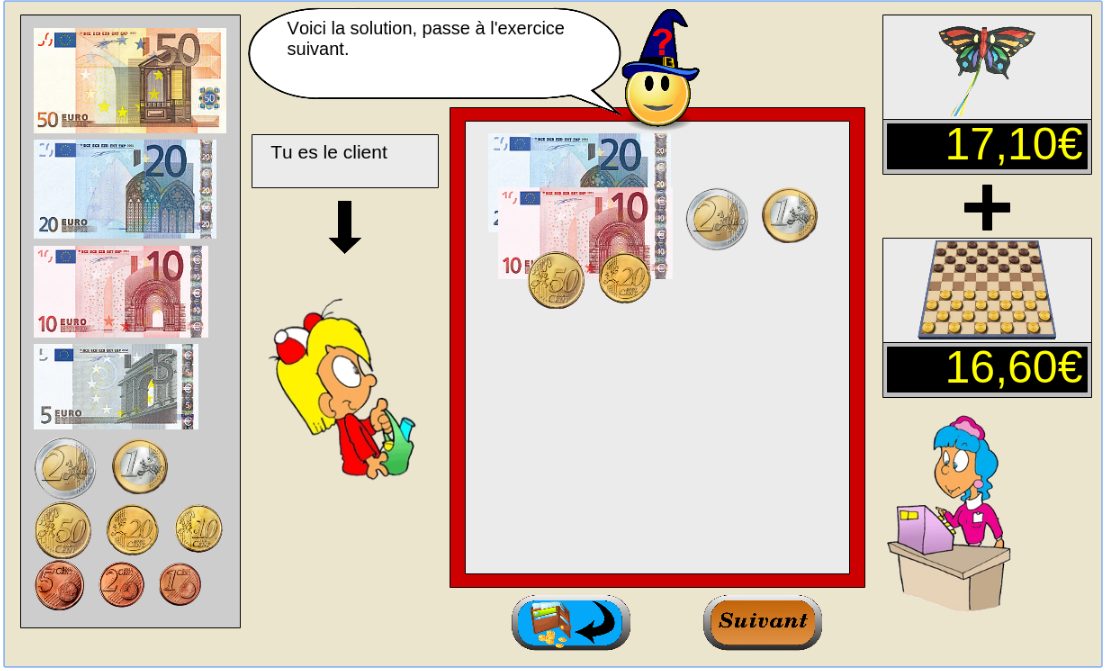}
    \includegraphics[width=0.45\columnwidth]{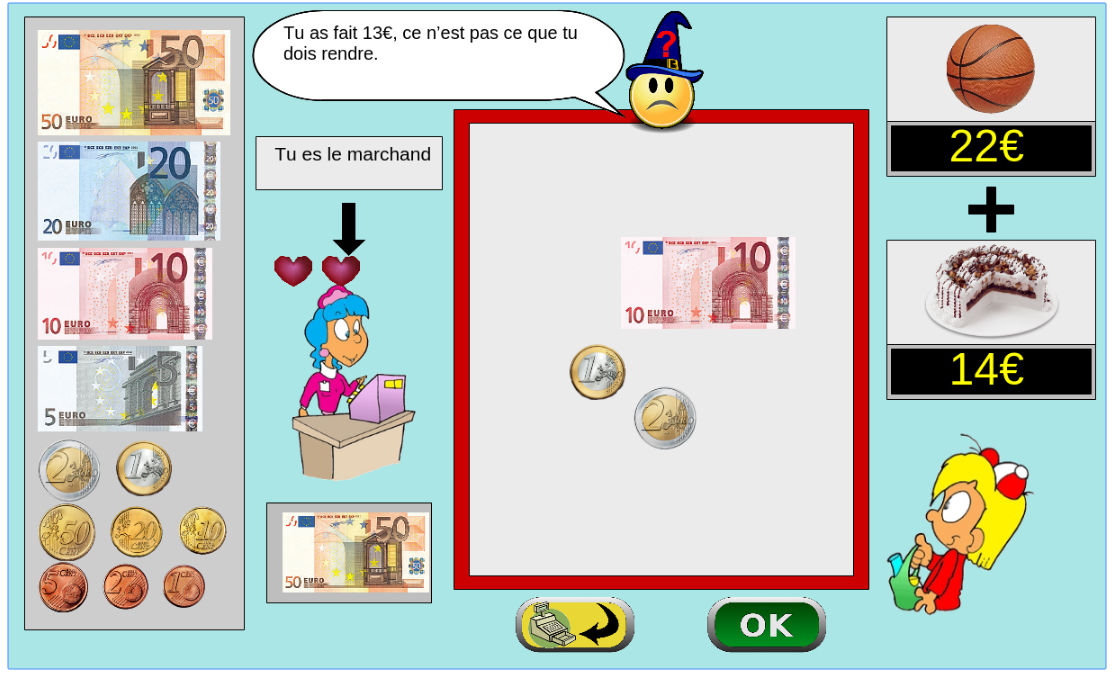}
  \caption{Four principal regions are defined in the graphical interface. The first is the wallet location where users can pick and drag the money items and drop them on the repository location to compose the correct price. The object and the price are present in the object location. Four different types of exercises exist: M : customer/one object, R : merchant/one object, MM : customer/two objects, RM : merchant/two objects. }
  \label{fig:monnaieinterface}
\end{figure}

This scenario is instantiated in a browser environment where students are proposed exercises in the form of money/token games (see Figure \ref{fig:monnaieinterface}). For an exercise type, one object is presented with a given tagged price and the learner has to choose which combination of bank notes, coins or abstract tokens need to be taken from the wallet to buy the object, with various constraints depending on exercises parameters. The seven Knowledge Components (KC) aimed at in these experiments are: a) \textbf{KnowMoney}: Global skill characterizing the capability to handle money to buy objects in an autonomous manner; b) \textbf{SumInteger}: Capability to add integers; c) \textbf{SubInteger}: Capability to subtract integers; d) \textbf{DecomposeInteger}: Capability to decompose integers into groups of ten and units; e) \textbf{SumCents:} Capability to add decimal numbers (cents); f) \textbf{SubCents:} Capability to subtract decimal numbers (cents); g) \textbf{DecomposeCents}: Capability to decompose decimal numbers (cents).

The various activities are parametrized in order to allow students to acquire a greater flexibility in using money. There are 11 parameters organized hierarchically. First, the \textbf{Exercise Type} is chosen : the student can be the costumer or the merchant and buy or give change with one or two objects. For each type of exercise the difficulty is chosen based on the \textbf{Difficulty} of decomposing a number. A number can be easy to decompose if there is a direct relation with a real bill/coin $a=(1,2,5)$ and hard to decompose if it requires more than one item $b=(3,4,6,7,8,9)$. The exercises will be generated by choosing prices with these properties and picking an object that is priced realistically. Another parameter controls the \textbf{Price Presentation}: in a written form and/or using a speech synthesizer. We allow to vary the \textbf{Cents Notation} due to the different practices in stores and countries that do not always follow the standardized rule. Finally we also consider the use of different \textbf{Representation of Money}: Real Euro or using poker tokens, that could reduce the visual ambiguity.

When the student begins the activity, one or two objects with their corresponding prices are shown. To complete the exercise the student has to drag and drop the money that it wants to use from the wallet location to the repository location. It is possible to request extra cues, by clicking on the face. To submit the answer it is necessary to click on the OK button. The feedback is then shown. If the answer is correct, the feedback is ``Congratulation you can move on to the next exercise''. We want to provide an experience that provides the most pedagogical gains and so, the student has 3 opportunities to solve the exercise and extra cues are provided each time the students makes a try. If after 3 trials the answer is still wrong a feedback with the correct solution is given and then the system goes to the next exercise.

In order to evaluate our algorithms, we use as baseline an optimized sequence created based on instructional design theory, whose reliability has been validated through several user studies, see \citeN{roy12math}. This baseline sequence grows in terms of complexity of the problem and simultaneously in terms of the difficulty of interaction. The prices produced, as seen before, become more complex in terms of the difficulty of decomposing a number and not on its absolute value. That is, the prices presented can be directly matched with the corresponding items, while the others require the composition of several items. Also the introduction of cents increases the complexity in several dimensions, requiring understanding of the concept of decimal and also on how to represent them. The introduction of tokens allows students to work with decimal numbers directly. Using cents is easier with real money as the items for integers (bills) and cents (coins) are different. The full details of this sequence are presented in Section~\ref{sec:PredefinedSeq}.

We do not use as baseline a random policy because this leads to too much errors, and changes on types of exercises that is disturbing for many of the students and not acceptable for the teachers.

\section{Intelligent Tutoring Systems with Multi-Armed Bandits}
\label{sec:LearningProblem}

To define an ITS we need to define a set of activities $A$ that the student can use to acquire these skills/knowledge components. If there is some knowledge about the domain, or the student's knowledge state, such information can be incorporated. The different algorithms we will propose vary in the amount of such expert knowledge that is required. The goal of an ITS system is, at each point in time, to propose students the activities most likely to increase their average competence level over all knowledge components based on previous students' performances. 

We will start this section by showing how multi-armed bandits can be used to optimize online teaching sequences as initially introduced by \cite{Clement2014edm}. We will then introduce two algorithms: ZPDES that does not use any explicit knowledge about the students relying only on the successes and failures on the exercises to base the choice of the next exercise; and RiARiT that explicitly estimates the level of the student's proficiency (using a process similar to bayesian knowledge tracing) to base its choice of exercises.

\subsection{Multi-Armed Bandits for Online Optimization of Teaching Sequences}
\label{sec:RiARiT}
To address the optimization challenge for ITS, we rely on state-of-the-art multi-arm bandit techniques (MAB)\cite{auer2003nonstochastic,bubeck2012MAB}. Following a casino analogy, multi-armed bandits describe the problem of finding the machine that provides the maximum reward,  initially unknown. To find the best machine we need to spend money exploring all of them before being able to bet always on the best one. This boils down to what is called the ``exploration/exploitation'' trade-off in machine learning, where we have to simultaneously try new activities to know which ones are the best, but also select the best ones so that the student actually learns. We here adapt such approaches to ITS \cite{Clement2014edm}, where the gambler is replaced by the teacher, the choice of machines is replaced by a choice of a learning activity, and reward is replaced by learning progress of the student (which is a proxy for maximizing acquired skills). We make the assumption that the activities that are currently estimated to provide a good learning gain, must be selected more often. Prior work showed that this assumption is true for many classes of problems \cite{Lopes12ssp} and is intrinsically motivating for people \cite{gottlieb2013information}.

A particularity here is that the reward (learning progress) is non-stationary, which requires specific mechanisms to track its evolution. Indeed, here a given exercise will stop providing reward, or learning progress, after the student reaches a certain competence level. Also we cannot assume that the rewards are i.i.d. (independent and identically distributed) as different students will have different preferences and many human factors, i.e. distraction, mistakes on using the system, create several spurious effects. Thus, we rely here on a variant of the EXP4 algorithm, proposed initially by \cite{auer2003nonstochastic} that considers a set of experts and chooses the actions based on the proposals of each expert. For our case, the experts are a set of variables that track how much reward each activity is providing \cite{Lopes12ssp}. 

More precisely, for each activity $\mathit{a}$ we define the quantity $w_a$ that tracks its recent rewards (correlate of learning progress). Each time that such activity is used, we update this value as follows $w_a \leftarrow \beta w_a + \eta r$, where $r$ is a reward that measures the benefit that activity $\mathit{a}$ gave to learning. $\beta$ and $\eta$ allow to define the tracking dynamics of this estimation. We will later propose several ways to compute this reward in a way that measures learning progress.

At any given time, we will select an activity $\mathit{a}$ proportionally to: $p_i = \tilde{w}_a (1-\gamma)+ \gamma \xi_u$, where $\tilde{w}_a$ are the normalized $w_a$ values to ensure a proper probability distribution, $\xi_u$ is a uniform distribution that ensures sufficient exploration of the activities and $\gamma$ is the exploration rate.

A pure selection based on the previous probabilities would allow exploring all possible activities $\mathit{a}$ but this has two drawbacks. It can create a bad effect of having too many changes in the type of exercises being proposed, and often jumping from too easy to too difficult exercises. It might not be possible to explore all the activities to estimate the learning progress that each one provides. All this can reduce the motivation and engagement of the students.  In order to ensure that students remain in challenging but possible to achieve areas we will limit exploration. Motivated by the zone of proximal development (ZPD), we allow an expert to specify an evolving ZPD based on previous results of the students. The use of the zone of proximal development will provide three advantages. Improve motivation as discussed before; further reduce the need of quantitative measures for the educational design expert; and provide a more predictive choice of activities. The implementation of these principles will be applied to both algorithms albeit with different details and expert knowledge.

In the following subsection we will introduce two algorithms that vary on the assumptions on the student learning that will lead to different ways to compute the reward. The resulting algorithms are shown in Alg.~\ref{alg:SSBandit}.

\subsection{ZPDES Algorithm: Zone of Proximal Development and Empirical Success}

We will start by presenting an algorithm that requires little domain/user knowledge. For this we will take two sources of inspiration: the \textbf{zone of proximal development} \cite{lee2005signifying} and the \textbf{empirical estimation of learning progress} \cite{oudeyer2007intrinsic}.

As discussed before, focusing teaching in activities that are providing more learning progress can act as a strong motivational cue \cite{gottlieb2013information}. Without neither a cognitive nor a student model, the only way to estimate learning is through the correctness of the answer of the student. We will thus compute the learning progress $r$ as follows:
\begin{equation}
r = \sum_{k=t-d/2}^t \frac{C_k}{d/2} - \sum_{k=t-d}^{t-d/2} \frac{C_k}{d -d/2}  
\label{eq:empprogress}
\end{equation}
where $C_k=1$ if the exercise at time $k$ was solved correctly. At time $t$, the equation compares, for the last $d$ samples, the success of the last $d/2$ samples with the $d/2$ previous samples, providing an empirical measure of how the success rate is increasing. This reward allows to compute a measure of the quality of each activity, measuring how much progress an activity has provided in a recent time window. We note that both extreme cases, when an activity is already acquired or when it is impossible to solve, will both have a reward of zero. Activities that are providing faster progress are assumed to be better than others with slower progress.

\begin{figure}[htbp]
	\centering
		\includegraphics[width=0.99\textwidth]{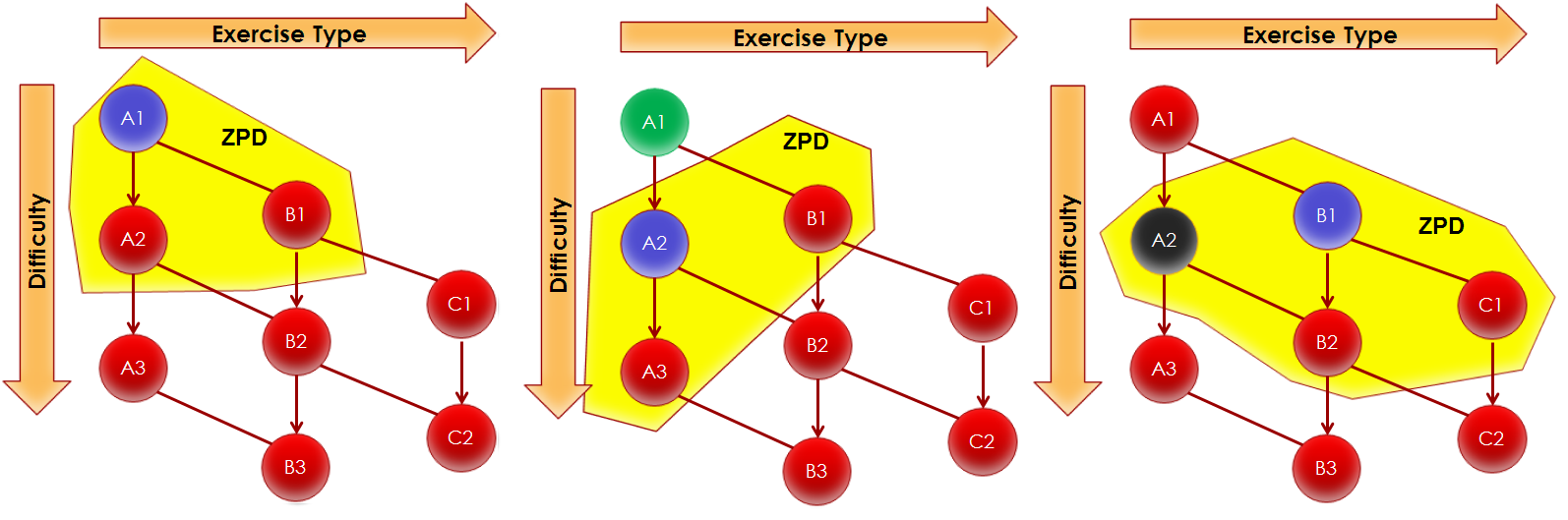}
	\caption{Example of the evolution of the zone-of-proximal development based on the empirical results of the student. The ZPD is the set of all activities that can be selected by the algorithm. The expert defines a set of pre-conditions between some of the activities ($A_1\rightarrow A_2\rightarrow A_3 \ldots$), and activities that are qualitatively equal ($A==B$). Upon successfully solving $A_1$ the ZPD is increased to include $A_3$. When $A_2$ does not achieve any progress, the ZPD is enlarged to include another exercise type $C$, not necessarily of higher or lower difficulty, e.g. using a different modality, and $A_3$ is temporarily removed from the ZPD. Both RiARiT and ZPDES make use of a ZPD mechanism but its definition and evolution is defined differently.}
	\label{fig:ZPDESgraph}
\end{figure}
 
Under this sole mechanism we still have too many activities to explore and we cannot rely on any knowledge about the level of the students to guide exploration. We allow an education expert to define the ZPD as a graph with the pre-conditions between activities. We also separate between subsets of activities that have a clear progression of difficulty, and other subsets of activities that might not have a clear progression of difficulty. For the scenario at hand the difficulty of the decomposition has a clear ordering while the price presentation and cents notation does not have a clear ordering. In practice to advance in the ZPD we proceed as follows. For activities at the same difficulty level we just allow a free exploration. For activities that have a clear progression in difficulty we will advance the ZPD based on the absolute success rate. 

Assuming a difficulty order of a subset of the activities $\mathit{a}_1<\mathit{a}_2<\ldots$, when the bandit level $w_a$ is below the level of the more complex parameter value, $w_{a_i}<\theta w_{a_{i+1}}$, with $\theta<1$ and the success rate is higher than a pre-defined threshold : $\sum_{k=1}^t \frac{C_k(j)}{t} > \omega$, we activate the parameter value $i+I$ with the following rule $w_{a_{i}}=0$ and $w_{a_{i+I}}=w_{a_{i+I-1}}$. The parameters $\theta$, $\omega$ and $I$ needs to be selected based on the desired variability of exercises.

The main intuition of this process is that when there are some activities whose difficulty grows, the ZPD will have to grow at the same rate. When activities do not have a clear order of difficulty, or that order might change from person to person, then it is necessary to allow wider exploration of the activities to accommodate individual differences.

\paragraph*{ZPDES algorithm} is very simple and uses very little domain knowledge. The expert teacher defines an exploration graph as in Fig.~\ref{fig:ZPDESgraph}. The simple use of the learning progress as a reward for each activity will allow to estimate the quality of each bandit. The algorithm proceeds as presented in Alg.~\ref{alg:SSBandit}.

\subsection{RiARiT Algorithm: Right Activity at the Right Time}

We propose another algorithm that is more informed about the domain and the student than what is used in the ZPDES algorithm. This extra information will be used to explicit estimate the knowledge level of the students and to compute a reward for the activities.

\paragraph*{Relation between KC and pedagogical activities}
\label{sec:RelationBetweenKCAndPedagogicalActivities}
In general, activities may differ along several dimensions and may take several forms (e.g. video lectures with questions at the end, or interactive games or exercises of various types). Each activity can provide an opportunity to acquire different skills/knowledge units, and may contribute differentially to the improvement over several KCs (e.g. one activity may help a lot in progressing in $KC_1$ and only little in $KC_2$). Vice versa, succeeding in an activity may require to leverage differentially various KCs. While certain regularities of this relation may exist across individuals, it will differ in detail for every student. Still, an ITS might use this relation in order to estimate the level of each student. Several approaches have been introduced to describe such relation between activities and KC, see \cite{desmarais2011performance} for a comparison.

Similar to a recent extension to Knowledge Tracing \cite{wang2013extending}, we model the competence level of a student in a given KC as a continuous number between $0$ and $1$ (e.g. $0$ means not acquired at all, $0.6$ means acquired at 60 percent, $1$ means entirely acquired). We denote $c_i$ the current estimate of this competence level for knowledge component $KC_i$. Then for each activity $\mathit{a}$ and $KC_i$ we define a value $q_i(\mathit{a}$) which encodes the competence level required in this $KC_i$ to have maximal success in this activity $\mathit{a}$. 


\paragraph*{Estimating the impact of activities over students' competence level in knowledge components}
\label{sec:EstimatingStudentsCompetenceLevel}

Key to our approach is the estimation of the impact of each activity over the student's competence level in each KC. This requires an estimation of the current competence level of the student for each $KC_i$. We do not want to introduce, outside activities, regular tests that would be specific to evaluate each $KC_i$ since it would have a high probability to negatively interfere with the learning experience of the student. Thus, competence levels need to be inferred through stealth assessment \cite{shute2008you,shute2011stealth} that uses indirect information coming from the combination of performances in activities and the $q$ values specified above.

The tracking of the competence levels $c_i$ could have been achieved using Knowledge Tracing \cite{corbett1994knowledge}. In our case we will rely on a simplified version based on the previously defined relation between activities and KCs. Let us consider a given knowledge component $i$ for which the student has an estimated competence level of $c_i$. When doing an activity $\mathit{a}$, the student can either succeed or fail. In the case of success, if the estimated competence level $c_i$ is lower than $q_i(\mathit{a})$, we are underestimating the competence level of the student in $KC_i$, and so should increase it. If the student fails and $q_i(\mathit{a})<c_i$, then we are overestimating the competence level of the student, and it should be decreased. For these two cases we can define a reward $r_i$ as:
\begin{equation}
r_i = q_i(\mathit{a})-c_i
\label{eq:rewardfromtable}
\end{equation}
Other cases provide little information, and thus $r_i=0$. We use this reward to update the estimated competence level of the student according to:
\begin{equation}
c_i =  c_i + \alpha r_i
\label{eq:updcomptlvl}
\end{equation}
where $\alpha$ is a tunable parameter that allows to adjust the confidence we have in each new piece of information. 

\paragraph*{Expert knowledge} can be incorporated as a set of global constraints on the ITS. Indeed, for example the expert knows that for most students it will be useless to propose exercises about decomposition of real numbers if they do not know how to add simple integers. Here the evolution of the ZPD can rely on explicit values of the estimated competence level of the student. Thus, the expert can specify minimal competence levels in given $KC_i$ that are required to allow the ITS to try a given activity $\mathit{a}$. Each activity is only explored if the student is already above this minimum threshold. We also allow the expert to define threshold for which a given activity is removed from the exploration.

\paragraph*{RiARiT algorithm} uses more information about the domain. The expert teacher defines a table with the relation between the activities and the KC, and also a set of minimum competence levels to activate a new activity. The relation between the success of an exercise, the estimated competence level and the required competence level of an exercise allows two things: a) to estimate the level of the student; and b) to compute a reward for that activity. The algorithm proceeds as presented in Alg.~\ref{alg:SSBandit}. The information required for this algorithm is more inline with other ITS systems. The knowledge required might be too difficult to give for an expert user when the number of activities, or KC, is high. Automatic  methods to fill such knowledge already exist and is an area of active research \cite{gonzalez2012dynamic,Baker2008more,gonzalez2014general,dhanani2014parameters}.

\begin{algorithm}
\caption{RiARiT and ZPDES ITS algorithms based on Multi-armed Bandits}
\label{alg:SSBandit}
\begin{algorithmic}[1]
\REQUIRE Set of $n_c$ Knowledge Componets $C$, set of $n_a$ activities $A$
\REQUIRE $\gamma$ rate of exploration
\REQUIRE distribution for parameter exploration $\xi_u$
\STATE Initialize $w_{a}$  uniformly
\IF{RiARiT}
\REQUIRE R Table
\STATE Initialize estimated competence levels $c^L$
\ENDIF
\WHILE{\textit{learning}}
\STATE Initialize ZPD
\STATE \COMMENT {Generate exercise:}  
    \FOR{$a \in ZPD$} 
    \STATE $\tilde{w}_a=\frac{w_a}{\sum_j w_j}$
    \STATE $p_a = \tilde{w}_a (1-\gamma)+ \gamma \xi_u$
    \STATE Sample $\mathit{a}$ proportional to $p_a$ 
    \ENDFOR
\STATE Propose activity $\mathit{a}$
\STATE Get student answer and compute reward
\IF{RiARiT}
\STATE Compute reward (Eq. \ref{eq:rewardfromtable})
\STATE Update competence levels (Eq. \ref{eq:updcomptlvl})
\STATE Update ZPD based on competence levels
\ENDIF
\IF{ZPDES}
\STATE Compute reward (Eq. \ref{eq:empprogress})
\STATE Update ZPD based on pre-requisites graph
\ENDIF
\STATE $w_a \leftarrow \beta w_a + \eta r$ \COMMENT{Update quality of activity}
\ENDWHILE
\end{algorithmic}
\end{algorithm}
%

\section{Simulations with Virtual Students}
\label{sec:Simulation}

We start by presenting a set of simulations to systematically test different properties of our algorithms. We define two different virtual populations of students to see how well the algorithm is able to select exercises adequate for each particular student and the impact of different properties of students. We will consider a population ``Q'' where all the students are able to use all the activities to learn, even if at different learning rates and with different maximum comprehension levels. Another population ``P'' aims at representing even more heterogeneous populations where each student might have a limitation for the comprehension of a particular type of activity. A concrete example is the case of a student that is not yet able to read will not be able to use exercises in written form to learn about mathematics, but if the exercise is presented in the spoken form it might be used for learning. Another example would be a student with hearing problems not able to solve an exercise that is presented in the oral form only.

We expect that in the population ``Q'' an optimization will not provide big gains because all students are able to use all exercises to progress. On the other hand, the population ``P'' will require that the algorithm finds a specific teaching sequence for each particular student.

Both models follow a standard Item Response Theory \cite{hambleton1991fundamentals}, where the probability of solving an exercise is given by:
\[
p(success) = \frac{\gamma(\mathit{a})}{1+e^{-(\beta(c^Q-c(\mathit{a})+\alpha))}}
\]
where $\beta$ and $\alpha$ are constants that allow to change the shape of the probability distribution and that can be chosen to provide different learning rates of a population. For model ``Q'' we have $\gamma(\mathit{a})=1$ meaning that all activities can be solved. For the model ``P'' some of the activities have $0\leq\gamma(\mathit{a})<<1$. This implies that for ``P'', some activities cannot be solved regardless of the competence level. The students have a probability of learning based on the difference between their levels and the level of the exercise.

\paragraph*{Results}
\label{sec:popsim}
We present here the results showing how fast and efficiently our algorithms estimate and propose exercises at the correct level of the students. Each experiment considers a population of 1000 students generated using the previous method and lets each student solve 100 exercises. For all populations the different initial, maximum final level of understanding of each KC is sampled from a truncated gaussian distribution. For the population ``P'' the values of parameter's understanding are sampled from four different distributions that include different levels of understanding ($\gamma$) for each parameter.	

\begin{figure}
    \centering
        \begin{tabular}{ccc}            
            &Q Students &P Students \\
            \rotatebox{90}{ Expert Sequence}&\includegraphics[width=.45\columnwidth]{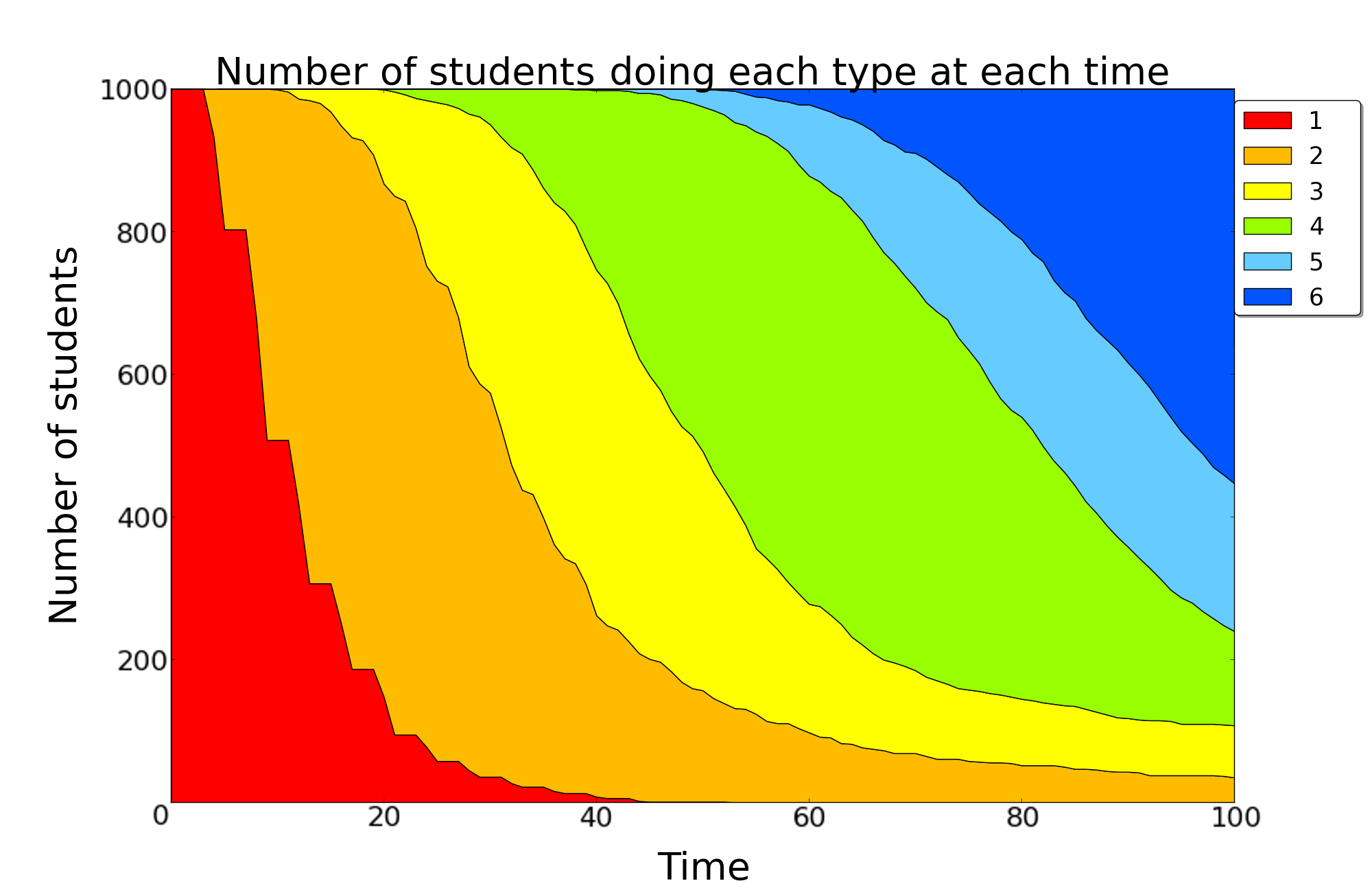}&
            \includegraphics[width=.45\columnwidth]{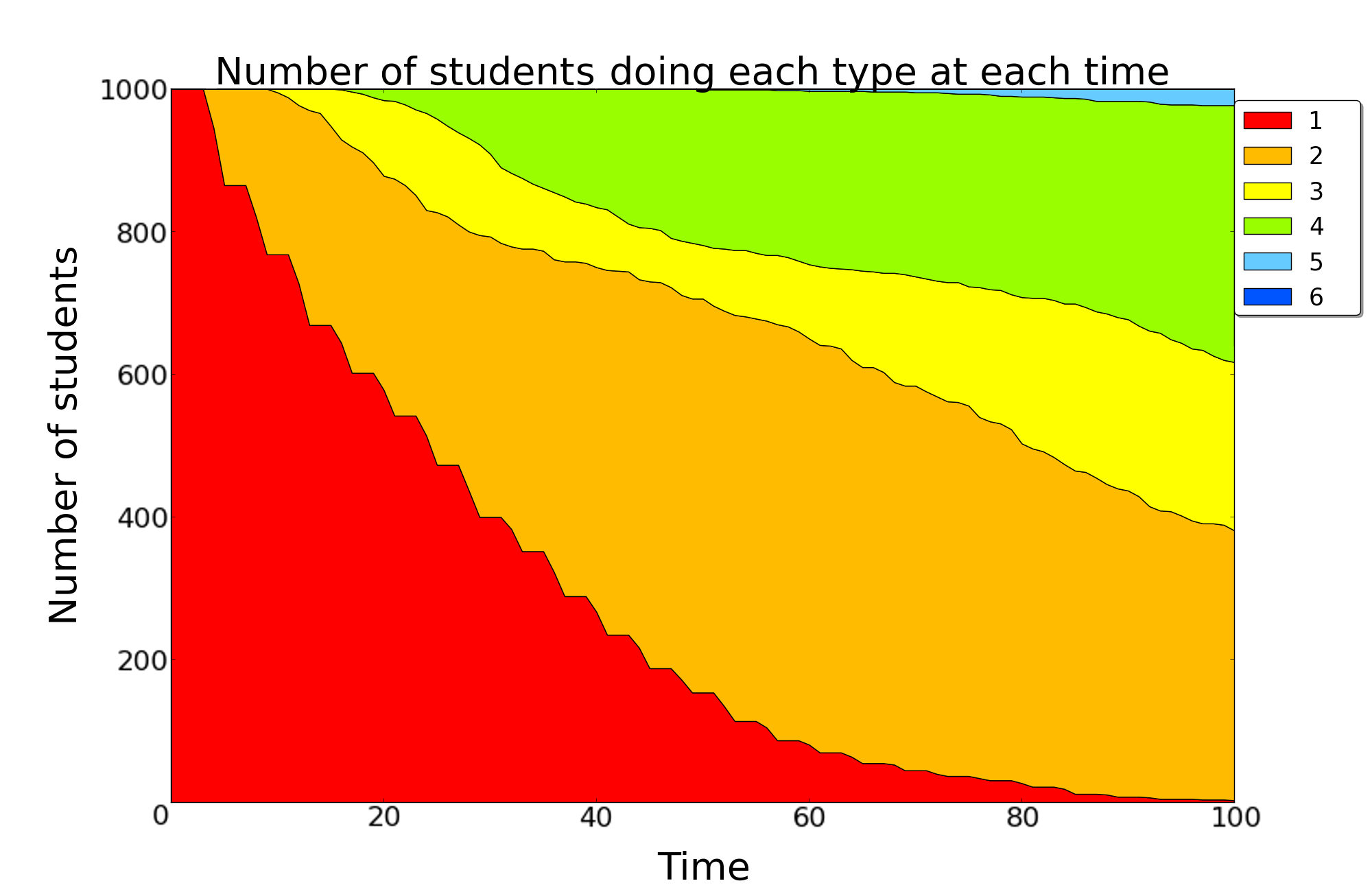}\\\hline
            \rotatebox{90}{  ZPDES}&\includegraphics[width=.45\columnwidth]{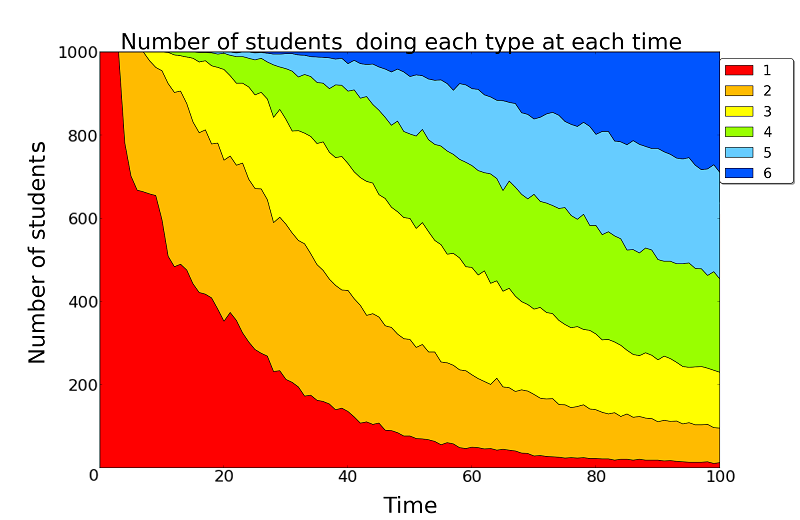}&
            \includegraphics[width=.45\columnwidth]{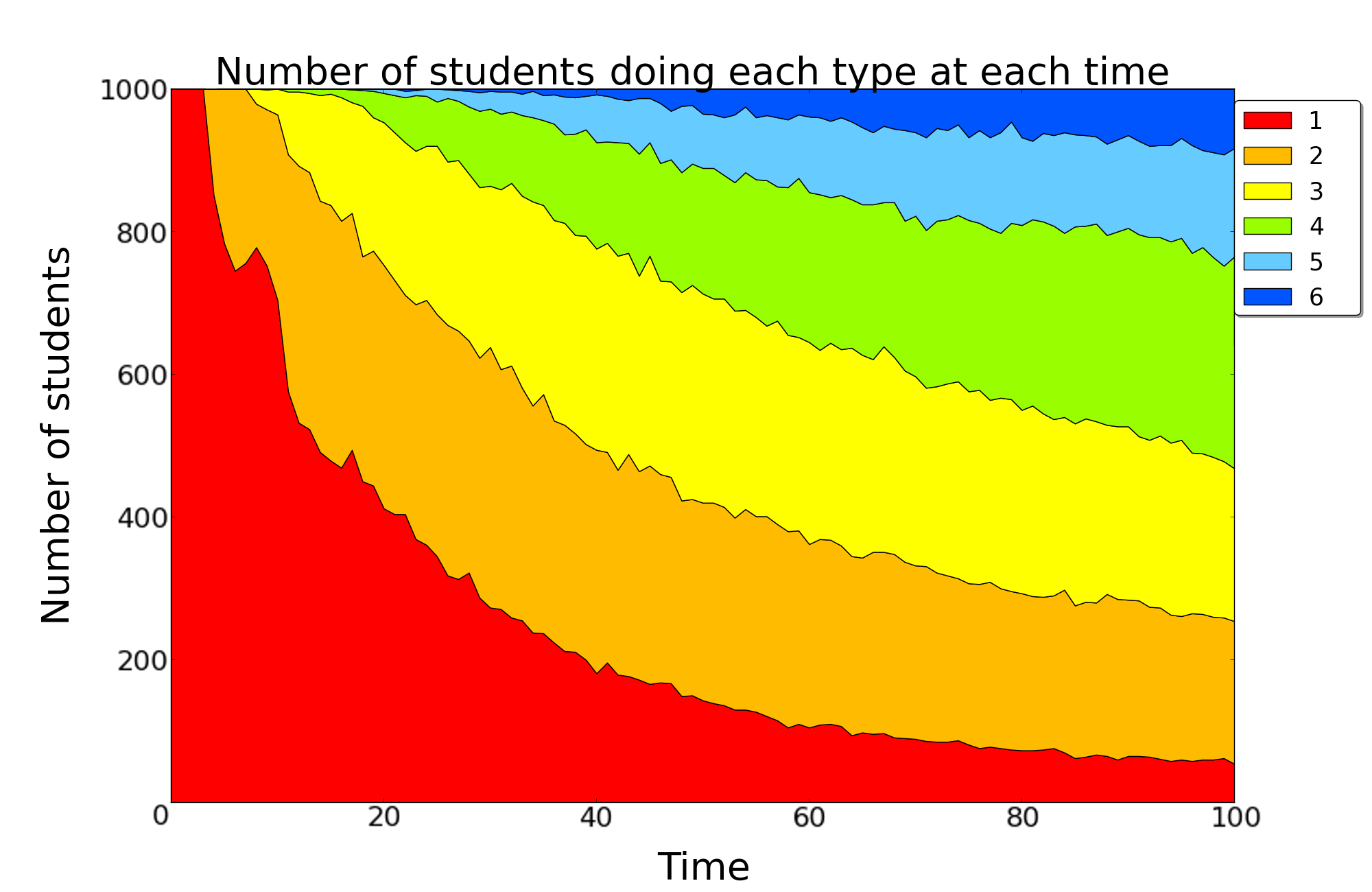}\\\hline
            \rotatebox{90}{  RiARiT}&\includegraphics[width=.45\columnwidth]{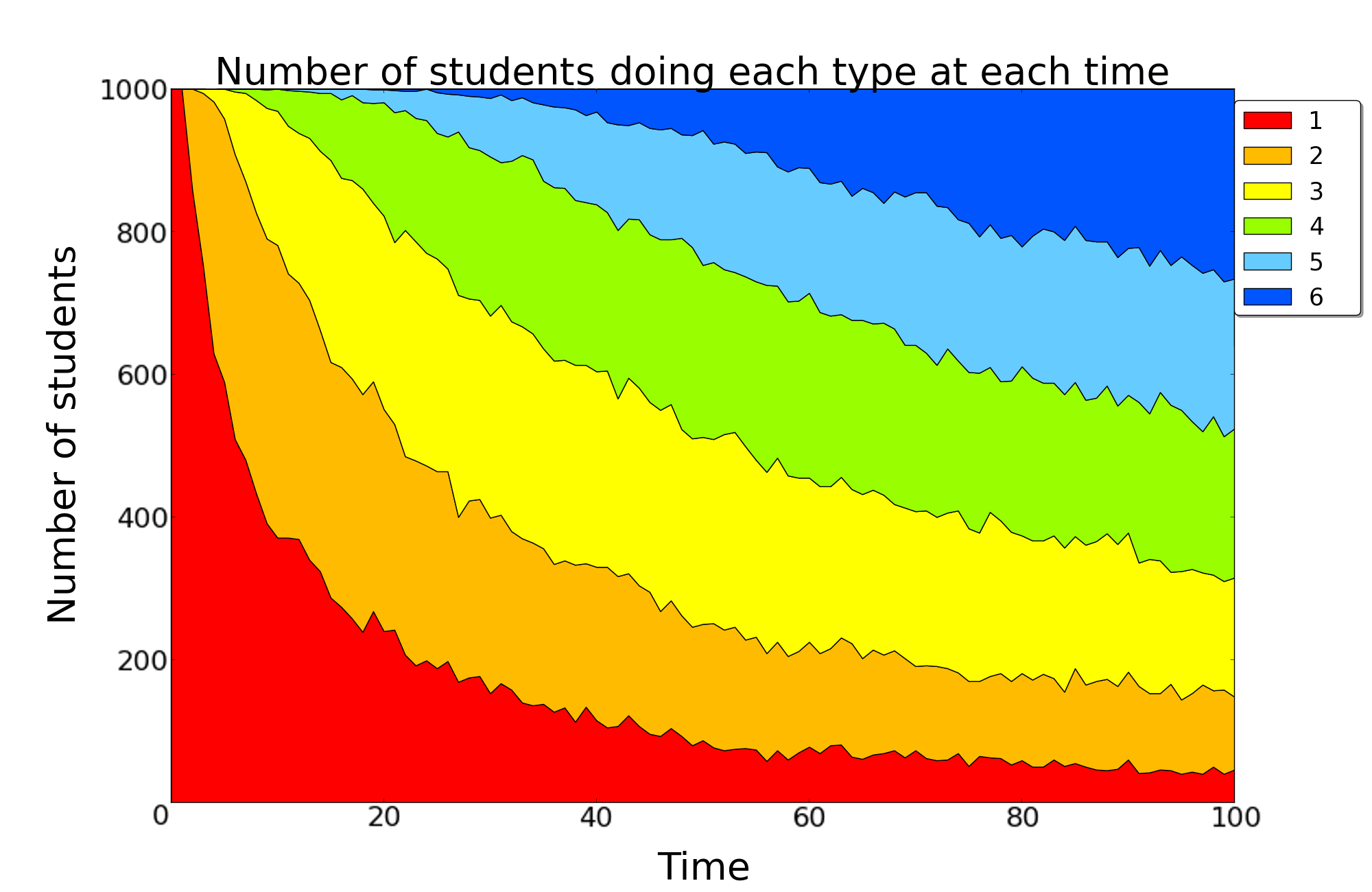}&
            \includegraphics[width=.45\columnwidth]{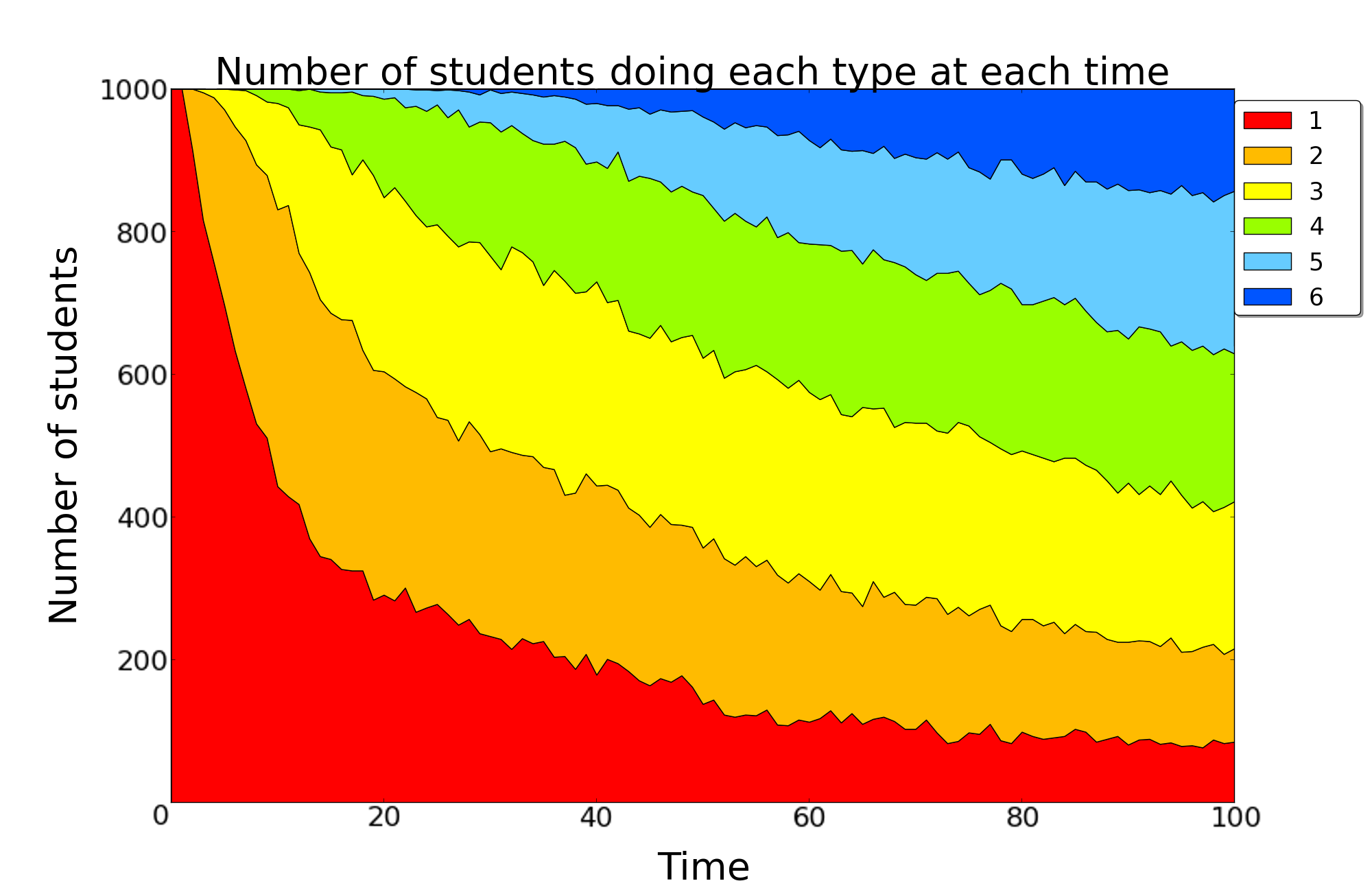}\\  
        \end{tabular}
    \caption{For each time instant, the curves show the total number of students being proposed each level of Exercise Difficulty. We can see that the Expert Sequence is not able to propose more difficult exercises as early. RiARiT and ZPDES can thus propose more difficult exercises sooner and keep proposing easier exercises longer. This shows the personalization properties of the algorithm.}
    \label{fig:Levels_time}
\end{figure}
\begin{figure}
    \centering
    \begin{tabular}{ccc}            
            &Q Students & P Students\\
            \rotatebox{90}{  Know Money}&
            \includegraphics[width=.45\columnwidth]{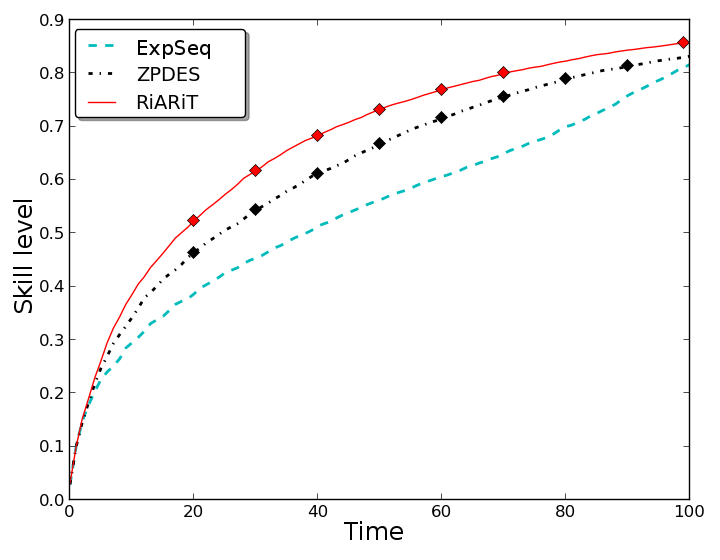}&
            \includegraphics[width=.45\columnwidth]{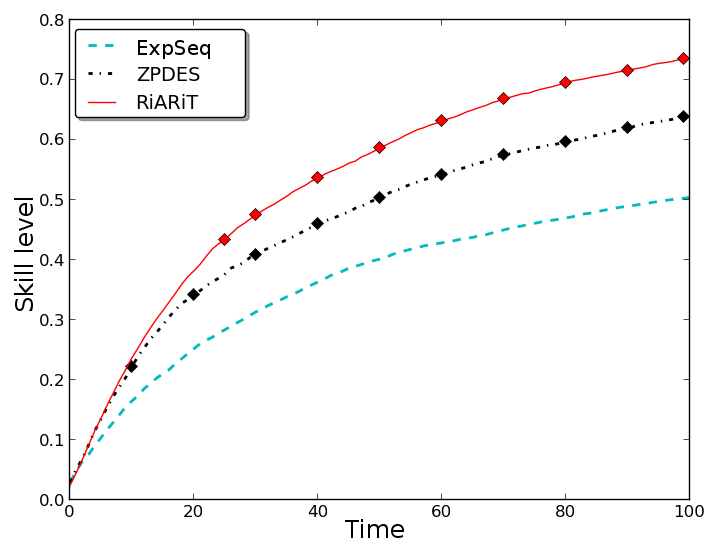}\\\hline
            \rotatebox{90}{  Integer Sum}&
            \includegraphics[width=.45\columnwidth]{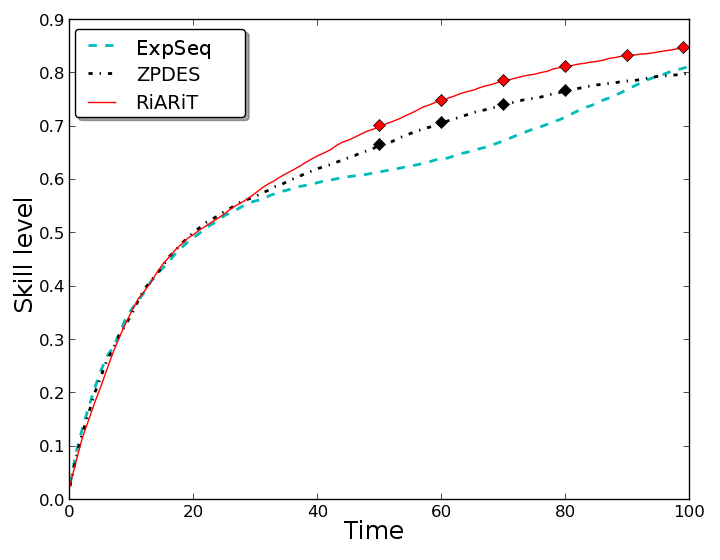}&
            \includegraphics[width=.45\columnwidth]{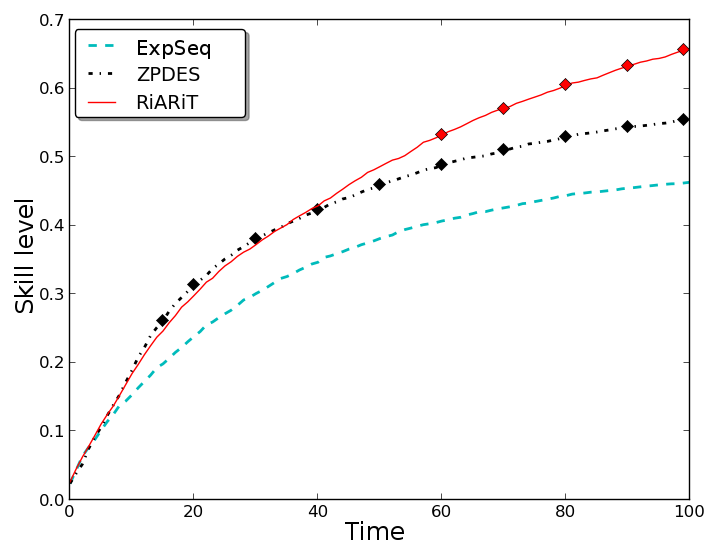}\\
        \end{tabular}       
    \caption{The evolution of the skill's levels of two KC with time for population ``Q'' and ``P''. Markers on the curve mean that the difference is statistical significant (red : RiARiT/ZPDES, black : ZPDES/ExpSeq). Both algorithms are able to improve upon the Expert Sequence.}
    \label{fig:level_evolAv}
\end{figure}

Figure \ref{fig:Levels_time} shows the number of students that are being proposed each type of exercise (only showing the parameter Difficulty for exercise Type M), independently if they succeed or fail the exercise. The actual student's levels are shown in Figure \ref{fig:level_evolAv}. We can see that in general, RiARiT and ZPDES start proposing more difficult exercises earlier while at the same time keep proposing the basic exercises much longer. This shows a clear adaptation to the actual level of the students.

Figure \ref{fig:level_evolAv} shows the skill's levels evolution during 100 steps. For Q students, learning with RiARiT and ZPDES is faster than with the expert sequence. For P students, as they might not understand particular activities, they block on certain stages due to the lack of adaptability of the expert sequence. On the other hand, ZDPES by estimating learning progress, and RiARiT, by considering the estimated level on all KC and parameter's impact, are able to propose better adapted exercises.
\begin{figure}
    \centering
      \begin{tabular}{cc}            
        Q Students &P Students \\
        \includegraphics[width=.49\columnwidth]{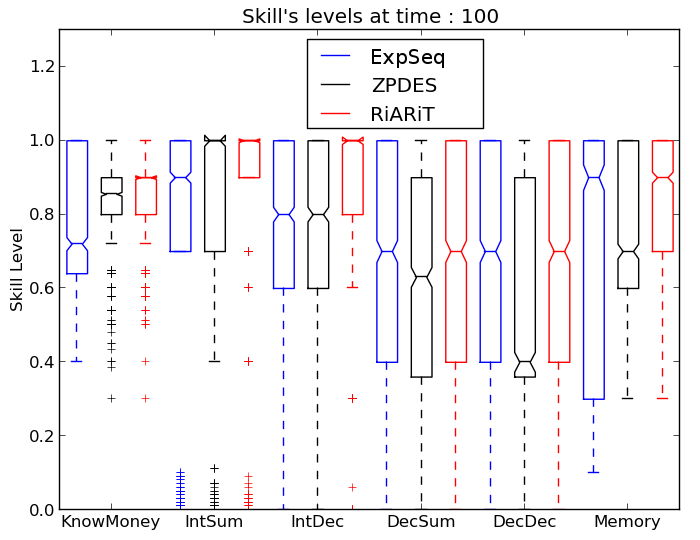}&
        \includegraphics[width=.49\columnwidth]{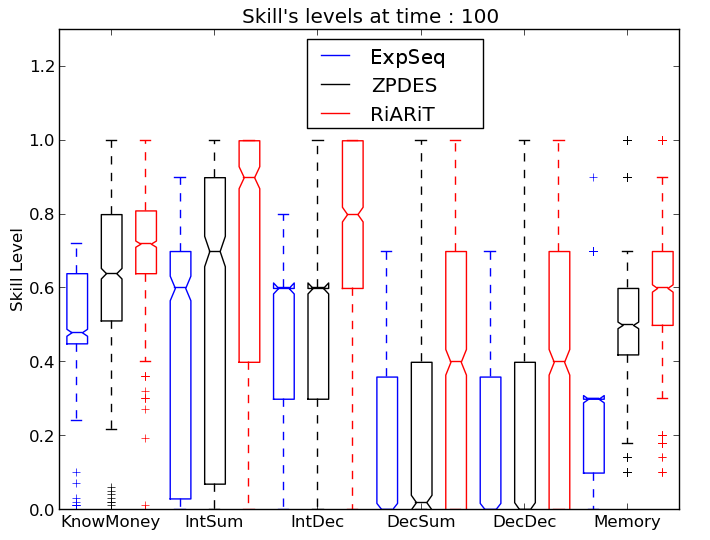}\\
      \end{tabular}
    \caption{Distribution of the acquired competence levels after 100 steps represented as a boxplot indicating median and the 4 quartiles. A statistical significant difference exists if the notches do not overlap. We can conclude that overall the automatic methods allowed a better understanding of all KC with a stronger gain in the case of P students.}
    \label{fig:mean_skills_level_sim}
\end{figure}

Figure \ref{fig:mean_skills_level_sim} shows the competence level of the students after 100 steps, represented as a standard boxplot. For ``Q'' and ``P'' students, differences are statistically significative for almost all KCs. RiARiT gives better results than Expert Sequence due to its greater adaptation to the students' levels. We can not distinguish between Expert Sequence and ZPDES. In the case of students of type ``P'', RiARiT and ZPDES are both better than the Expert Sequence This is explained because when the students are not able to understand a specific activity, an hand-designed sequence can not adapt to all possible variants of the students' learning.
\begin{figure}
    \centering
        \begin{tabular}{ccc}            
            &Q Students&P Students\\
            \rotatebox{90}{Expert sequence}&\includegraphics[width=.45\columnwidth]{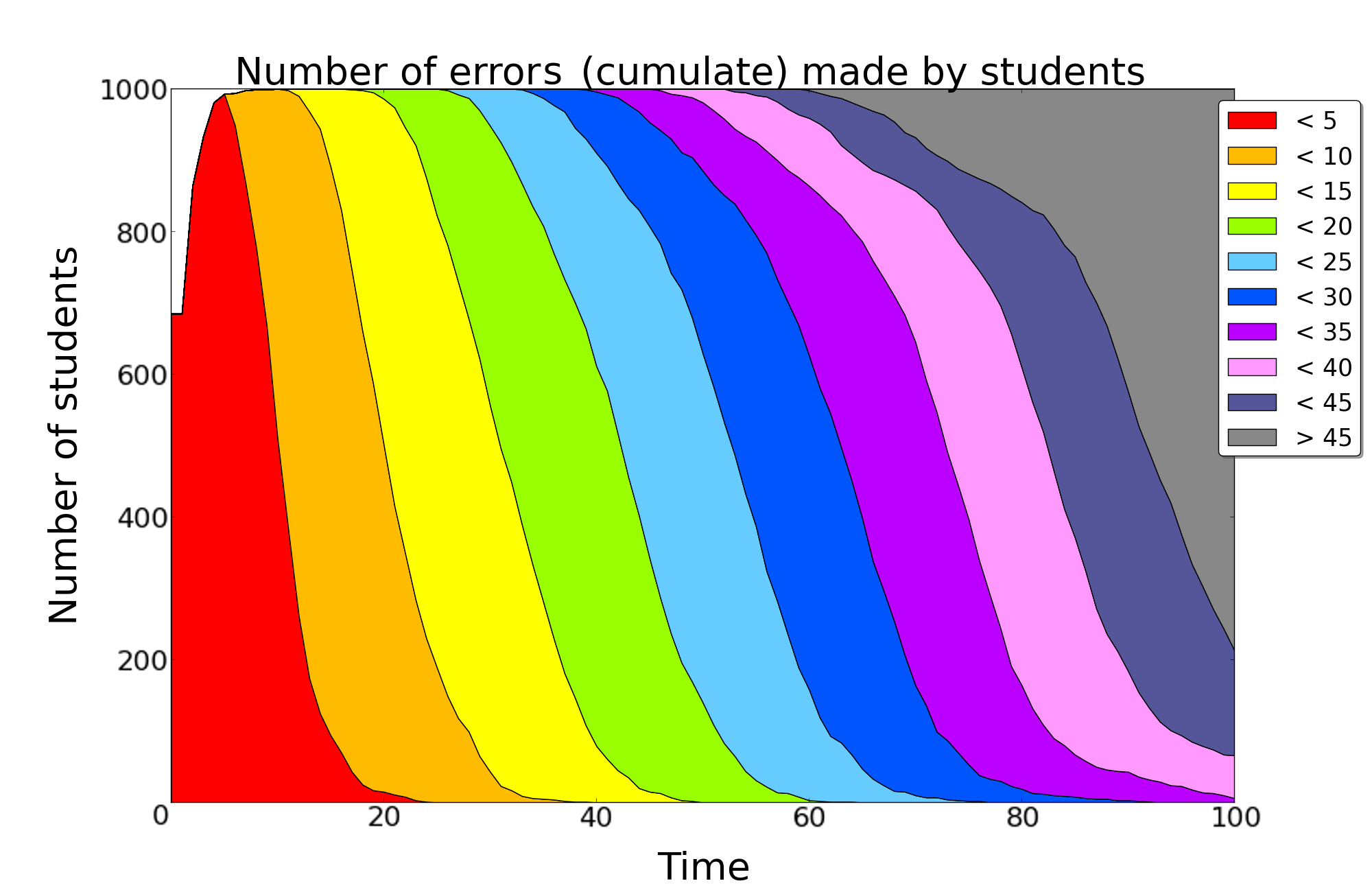}&
            \includegraphics[width=.45\columnwidth]{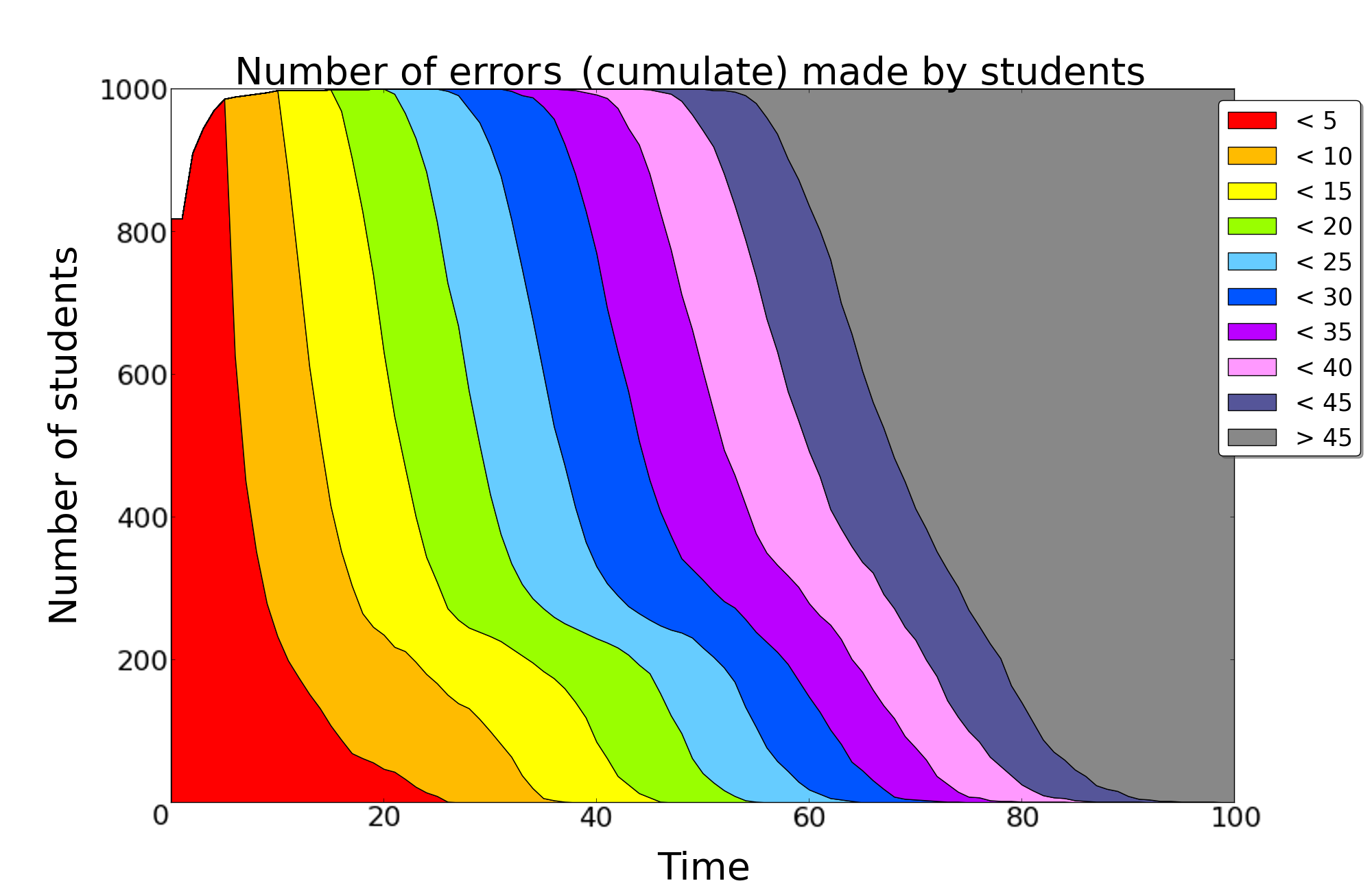}\\\hline
            \rotatebox{90}{  ZPDES}&\includegraphics[width=.45\columnwidth]{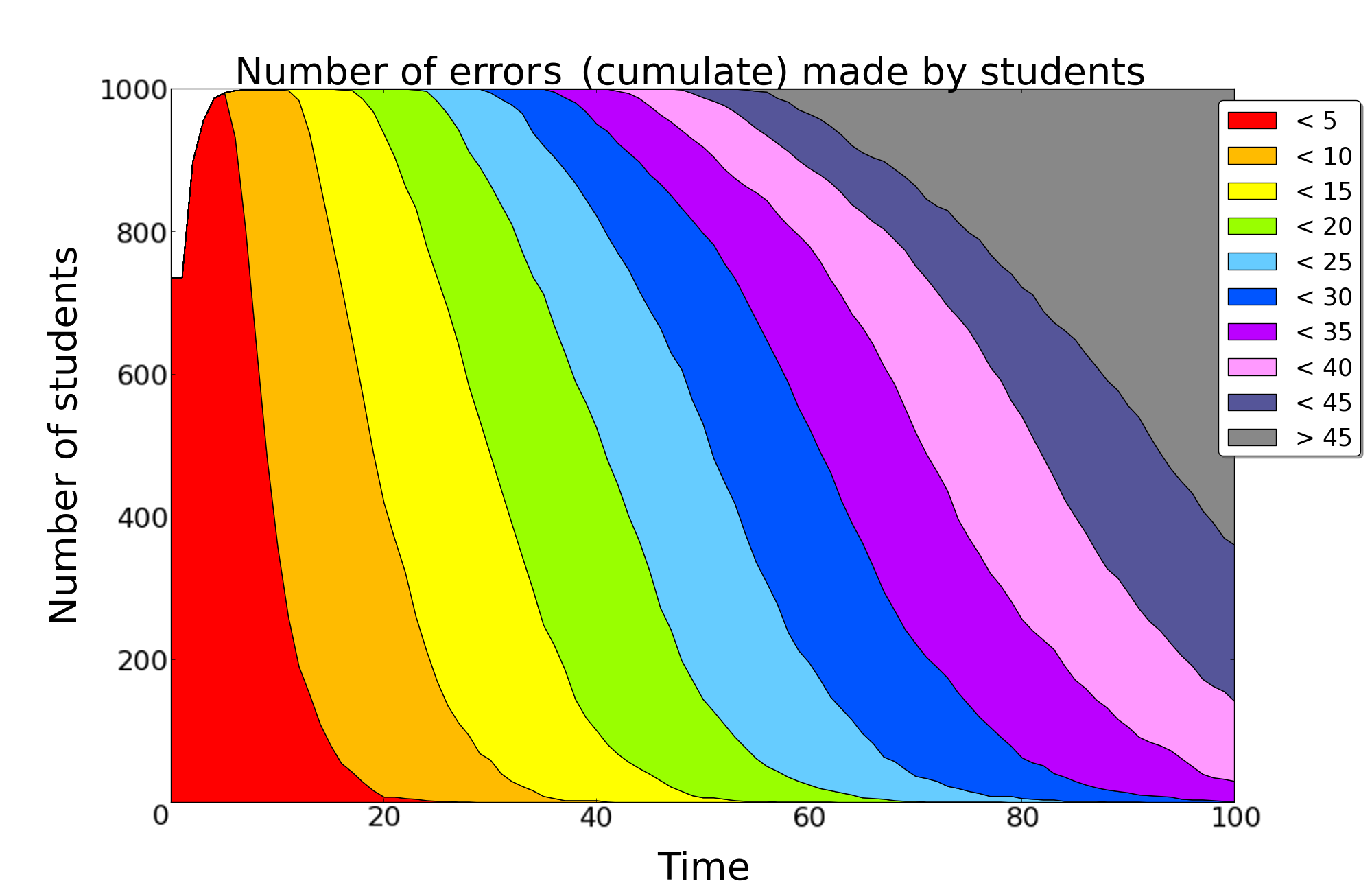}&
            \includegraphics[width=.45\columnwidth]{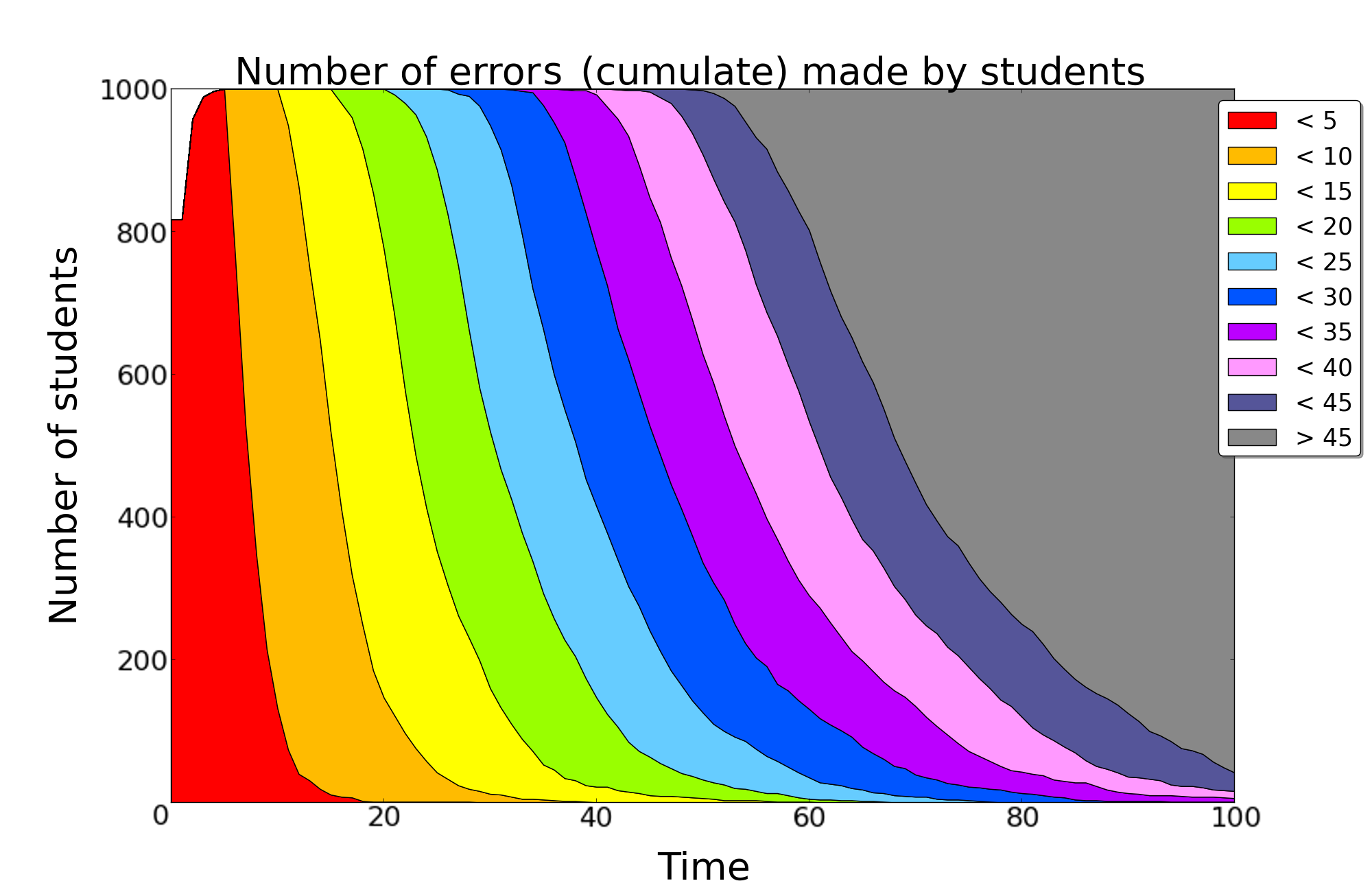}\\\hline
            \rotatebox{90}{  RiARiT}&\includegraphics[width=.45\columnwidth]{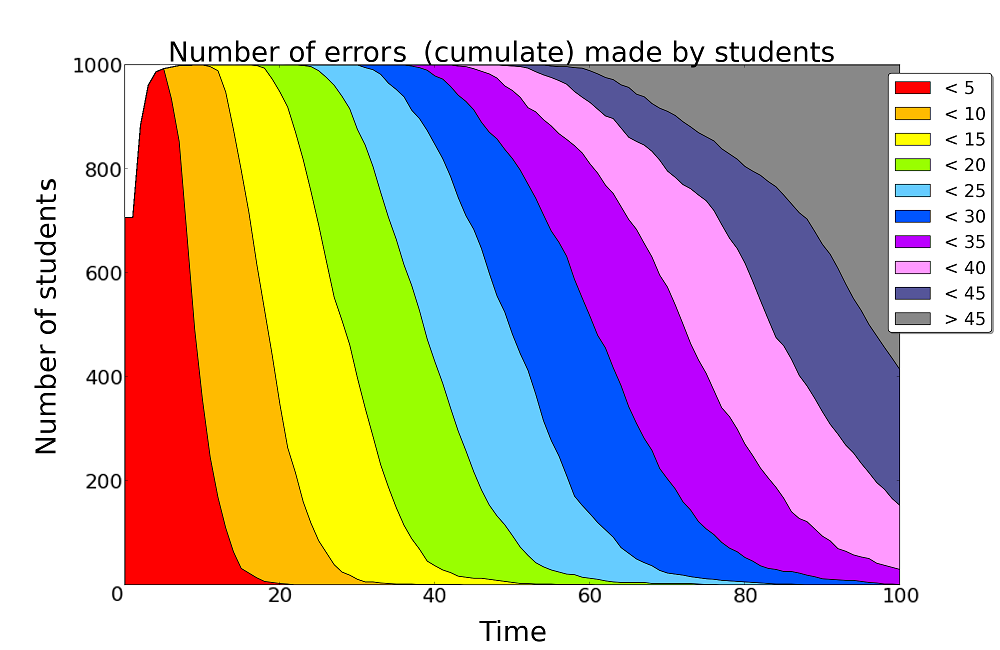}&
            \includegraphics[width=.45\columnwidth]{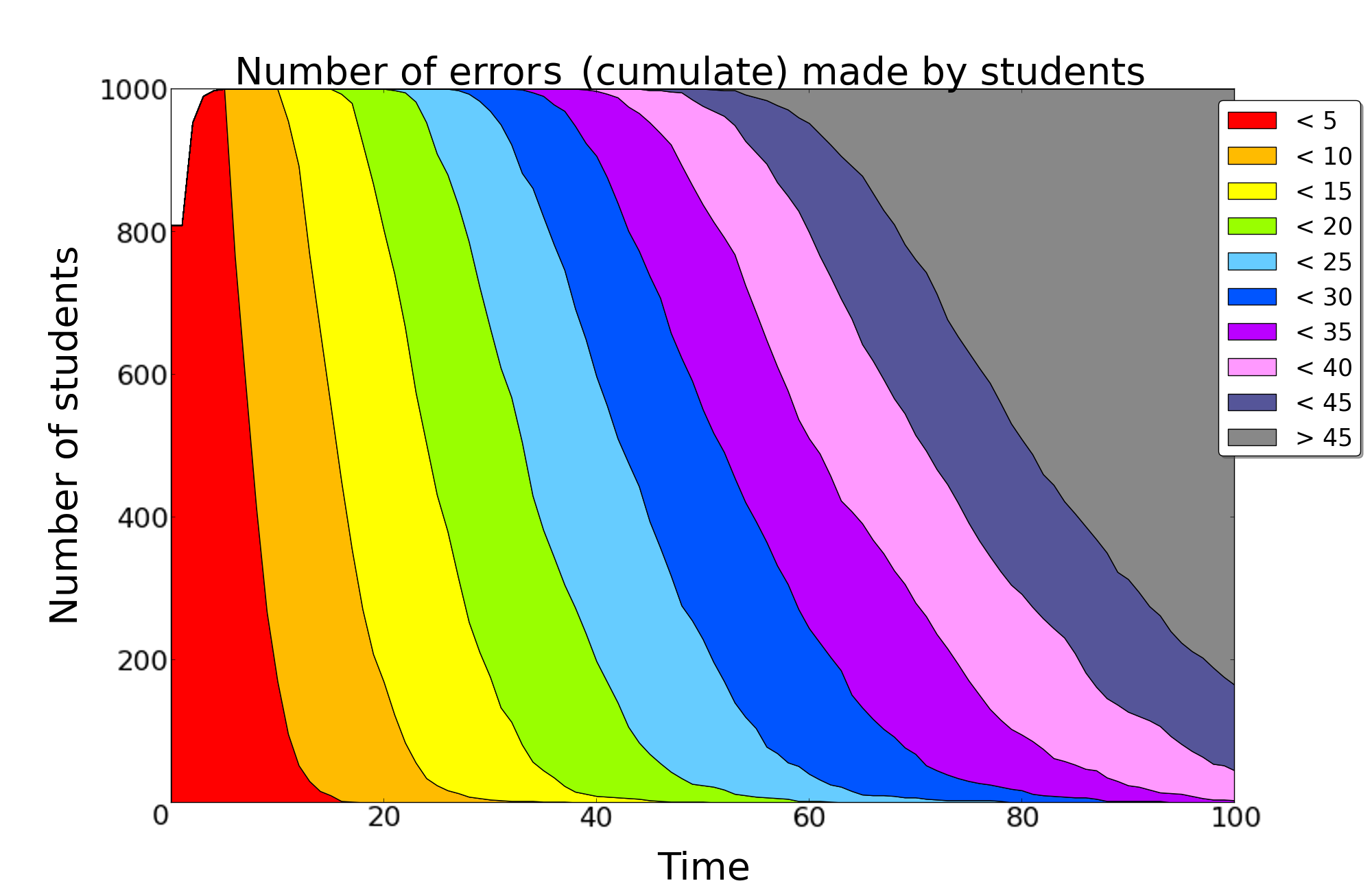}\\
        \end{tabular}
    \caption{This figure shows the number of errors made by the students. For each time instant, and for each number of cumulative errors (indicated in the colors), the curves shows the total number of students that made that number of errors.}
    \label{fig:cumulate_error}
\end{figure}

We can also analyze the errors that the students make during learning. If the exercises are too difficult to solve there will be many errors and this can be a source of frustration. Figure \ref{fig:cumulate_error} shows that for both types of students, at the beginning, the number of errors is equal among methods but with time, expert sequence gives rise to more errors than when using RiARiT or ZPDES, in particular for ``P'' students. And for ``P'' simulation, students have less errors with RiARiT than with ZPDES, showing that RiARiT has a better adaptation than ZPDES. 

\section{User Studies}
\label{sec:ResultsOnUserStudies}


As the final goal of an ITS is to provide a more efficient teaching experience to students, we performed a user study aiming at validating the software infrastructure, the interface and the algorithms themselves \footnote{The software is available at \url{https://github.com/flowersteam/kidlearn_lib}}. We want to evaluate principally the learning improvement, the personalization, and the impact of the use of a model. We considered $11$ different schools  in the Bordeaux metropolitan area. We had a total of $400$ students between $7$ and $8$ years old. We divided each class into 4 groups, with one control group where student do not use the software and 3 groups where exercises are proposed using : a)  Expert Sequence; b) ZPDES; c) RiARiT. To measure student learning, students pass a pre-test a few days before using the ITS, and a post-test a few days after using the ITS. The control group pass the pre- and post-test at similar time frame but without using the game.

For this experiment, and due to constraints of the schools, students had 40 minutes to do the exercises. Each student completes a different number of exercises. This makes the comparison of results between the different algorithms harder but, on the other hand, it is a real constraint when using class time. In the following results, the axis ''Time'' represents the succession of exercises. For example, ''Time 1'' is the first exercise for all students, but if at time 30, some students have already finished, they don't do the exercises at time 30, so with time the cumulative number of students decreases.

\subsubsection*{Maximum level achieved}
\begin{figure}
    \centering
        \begin{tabular}{cc}
            Maximum level reached & Maximum level achieved \\
            \includegraphics[width=.49\textwidth]{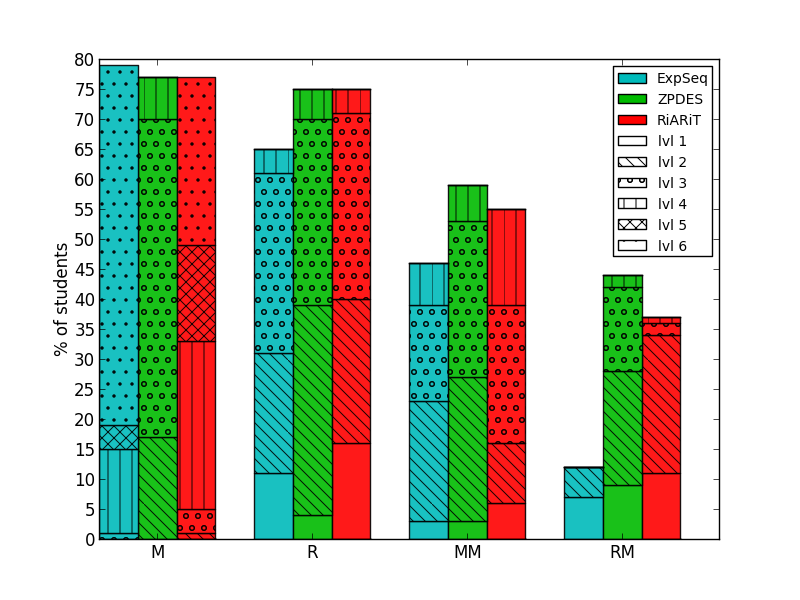}&
            \includegraphics[width=.49\textwidth]{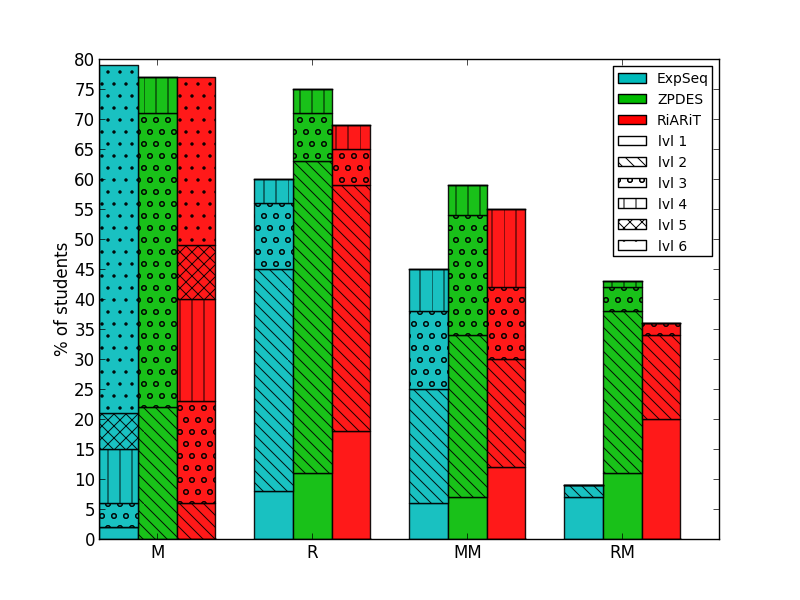}\\  
        \end{tabular}
    \caption{The figures show the proportion of highest level reached (left) or achieved (right).  A level can be reached, yet not achieved. We can see that ZPDES and RIARIT allowed students to reached and succeed the most challenging types of exercises (MM and RM). By combining this with the information from Fig. \ref{fig:Levels_time} we can see that students are reaching their level of competence earlier when using the automatic algorithms.
}
    \label{fig:maxReachAchiev}
\end{figure}

%
%
Figure \ref{fig:maxReachAchiev} shows the percentage of students who succeeded each level and type of exercise. The graphic is not cumulative, so students are taken into account only for the maximum level they reach for each type of exercise. Globally, we can see that there is much more students who succeed higher levels of R, MM and RM exercises with the RiARiT and ZPDES algorithms than with the Expert Sequence. To know if the type of sequence management have a significant impact on the maximum level succeeded by students, we did a $\chi^2$ test to test the dependence and an \textit{ANOVA} to test if the differences are significant. Tests results have been summarized in Table~\ref{tab:achiStatsResults}. The first part shows the student medium level for each group, we can see that students have succeeded highest level exercises with ZPDES and RiARiT than with the Expert sequence except for M type. This is not surprising as the M exercises are the first ones to be proposed and the Expert Sequence spent more time there. The second part of the table shows the p-value of $\chi^2$ test for independence. We can see that, for the majority of exercises type (M,R,RM), the p-value is lower than 5\%, so we can reject the null hypothesis of independence. Then to improve our analysis, we also did an \textit{ANOVA} to ensure that the differences between groups are significant. We can see that, in majority, the \textit{ANOVA} allow to say that the differences are significant.

So even if there are much more students who reach and succeed the highest exercise of M type with the Expert Sequence (75\% versus 0\% for ZPDES and 35\% for RiARiT), there is much more students who reach and achieve the other types. ZPDES and RiARiT proposed exercises of other types that, in the end, results in a better acquisition of the KCs. For R type exercise : 95\% for ZPDES and 90\% for RiARiT of students succeed at least one exercise versus 75\% for Expert Seq. And the difference increase with MM and RM exercises.

\begin{table*}
    \centering
    \caption{Statistical test on the results of the user studies. The top table shows the average difficulty level reached for each type of exercise. Then we present two statistical test to verify if the difference in the means and in the distributions are statistically significant. From the results we can conclude that for most cases ZPDES is better than the Expert Sequence.}
    \label{tab:achiStatsResults}		
          \begin{tabular}{|c|c||c||c||c|}\hline
                          & \multicolumn{4}{c|}{level average}   \\\hline
                          &   M     & R       &   MM     & RM    \\\hline
          Expert      &{\bf5.42}& 1.66    & 1.41     & 0.14  \\\hline
          RiARiT          & 4.47    & 1.74  	  & 1.77     & 0.70  \\\hline
          ZPDES           & 2.79    & \bf2.01 & \bf1.83  &\bf1.05\\\hline
          \end{tabular}

          \vspace{0.5cm}
          \begin{tabular}{|c|c||c||c||c||c|}\hline
                            &                 \multicolumn{5}{c|}{test $\chi^2$ : p-values}                \\\hline
                            &   M             & R               &   MM   & RM             & All type       \\\hline
          Expert/RiARiT    & \bf$\ll$ .001  & \bf.04  & .17  & \bf$\ll$ .001 & \bf$\ll$ .001 \\\hline
          RiARiT/ZPDES    & \bf$\ll$ .001  & .14  & .059  & \bf$\ll$ .001 & \bf$\ll$ .001 \\\hline
          ZPDES/Expert    & \bf$\ll$ .001  & \bf$\ll$ .001  & .085 & \bf$\ll$ .001 & \bf$\ll$ .001 \\\hline
          \end{tabular}
          \vspace{0.5cm}
          \begin{tabular}{|c|c||c||c||c|}\hline
                            &                 \multicolumn{4}{c|}{ANOVA : p-values}                     \\\hline
                            &   M             & R        &   MM     & RM              \\\hline 
          Expert/RiARiT    & \bf$\ll$ .001  & .88     & .11     & \bf$\ll$ .001 \\\hline 
          RiARiT/ZPDES      & \bf$\ll$ .001  & \bf .04   & .70     & \bf$\ll$ .001 \\\hline 
          ZPDES/Expert     & \bf$\ll$ .001  & .07      & \bf .04   & \bf$\ll$ .001 \\\hline 
          \end{tabular}    
\end{table*}
%

\subsubsection*{Personalized Learning Sequences}
\begin{figure}
  \centering
        \begin{tabular}{ccc}
            \multicolumn{3}{c}{\includegraphics[width=.7\textwidth]{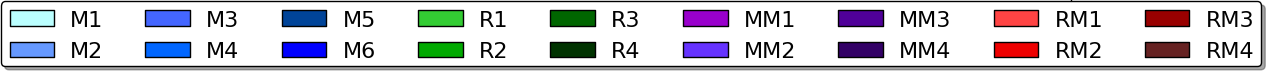}}\\
            \rotatebox{90}{\vspace{2.5cm} \bf Expert Sequence}&
            \raisebox{-.0\height}{\includegraphics[width=.55\textwidth]{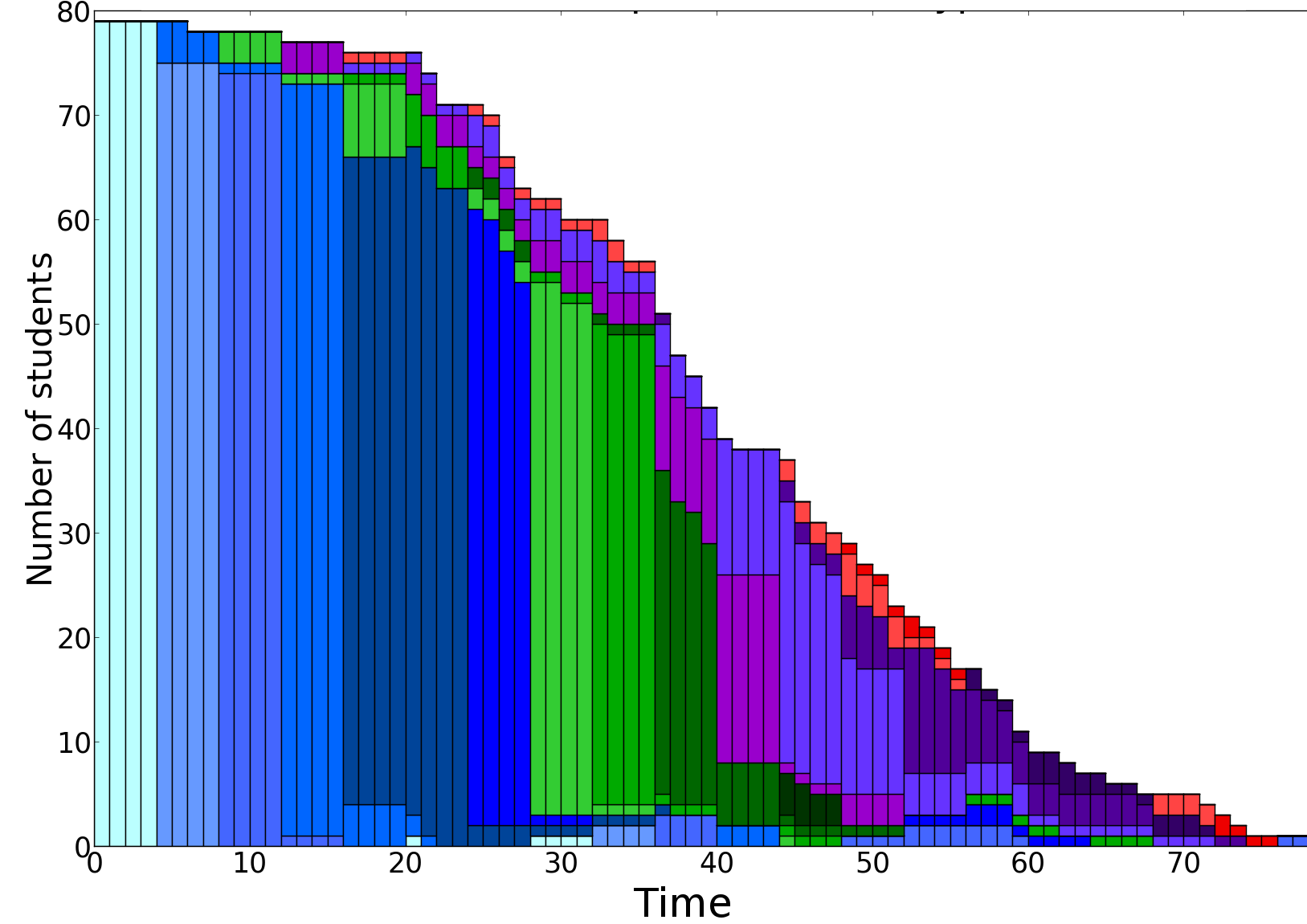}}&
            \includegraphics[width=.40\textwidth]{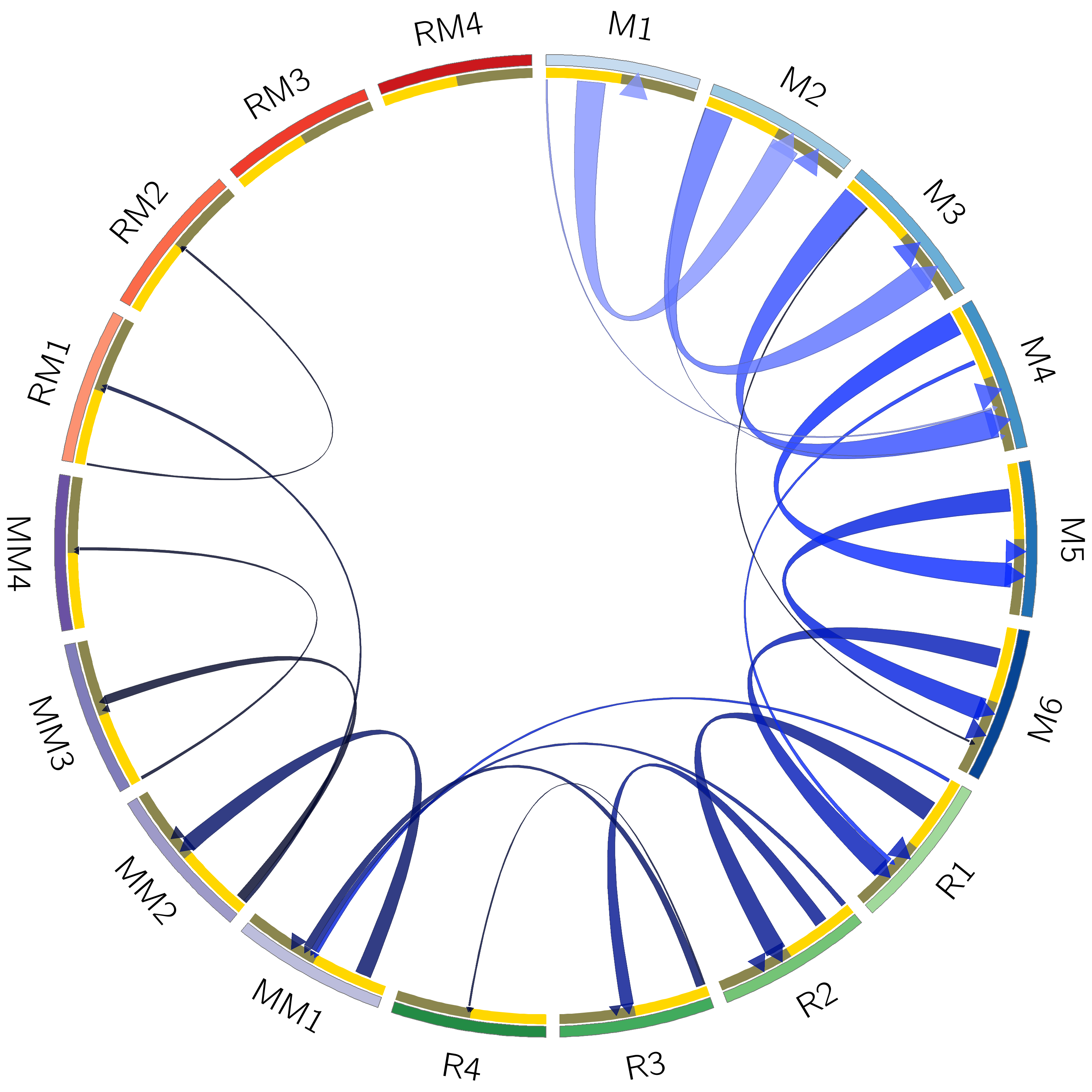}\\
            \rotatebox{90}{\bf ZPDES} &
            \raisebox{-.0\height}{\includegraphics[width=.55\textwidth]{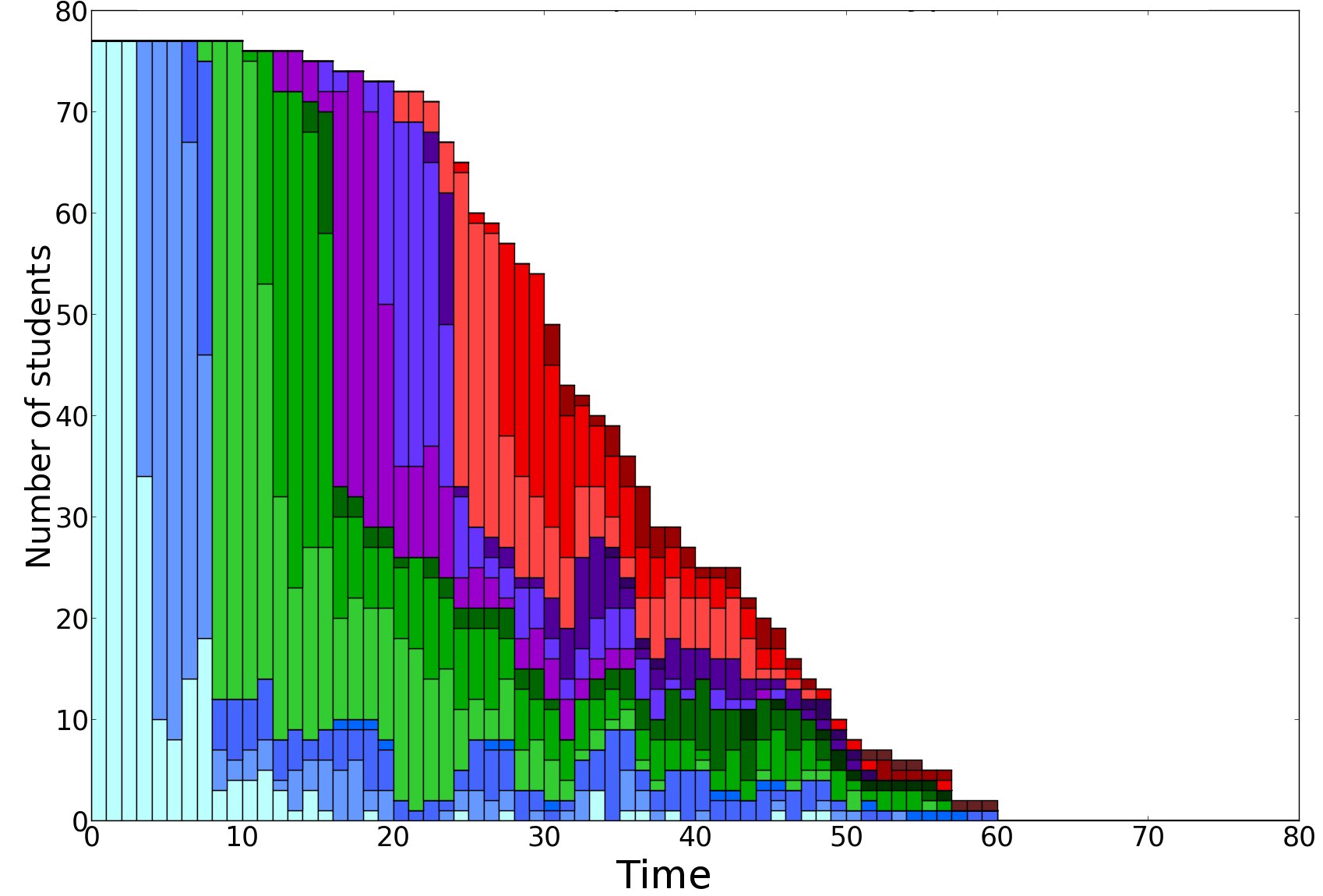}}&
            \includegraphics[width=.40\textwidth]{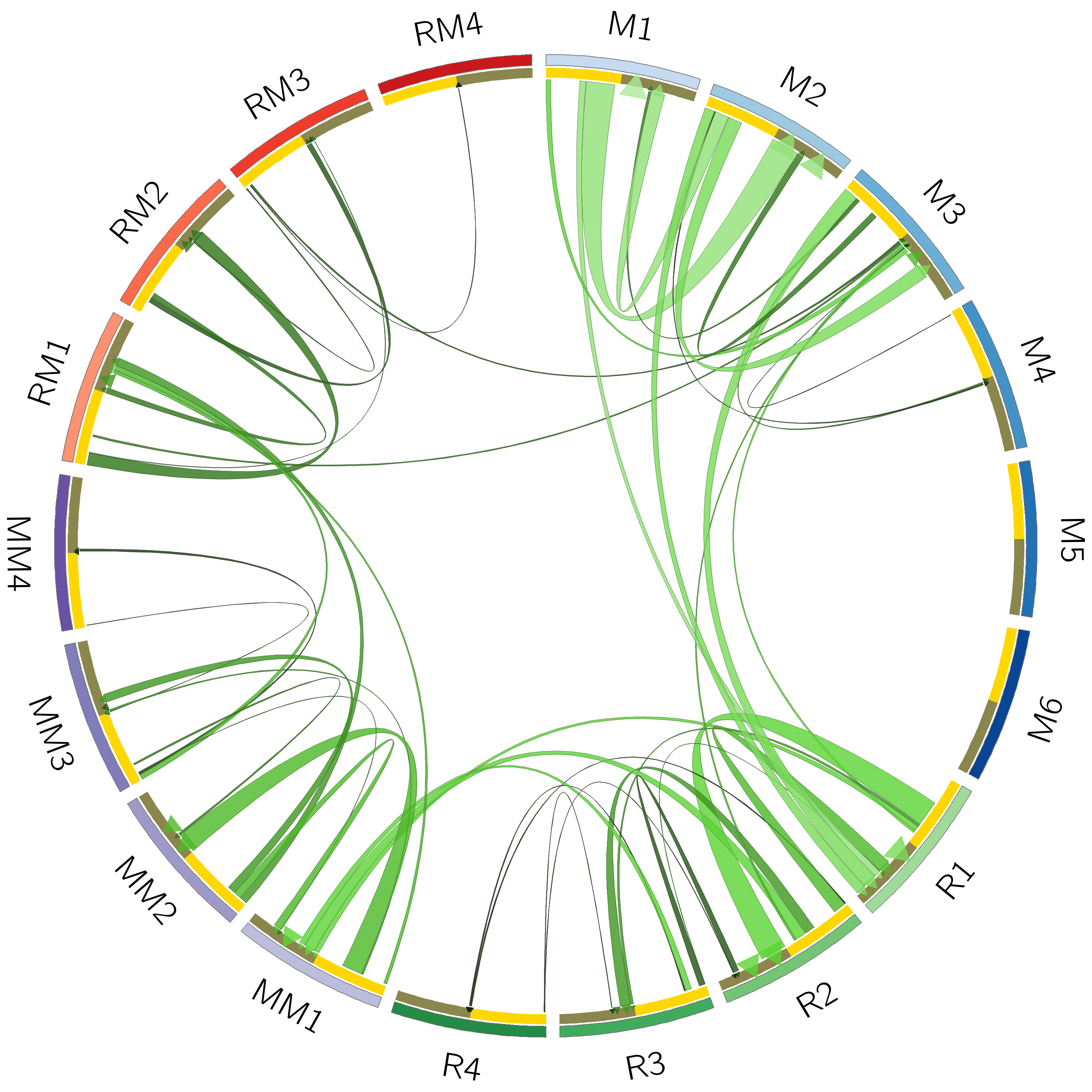}\\
            \rotatebox{90}{\bf RiARiT} &
            \raisebox{-.0\height}{\includegraphics[width=.55\textwidth]{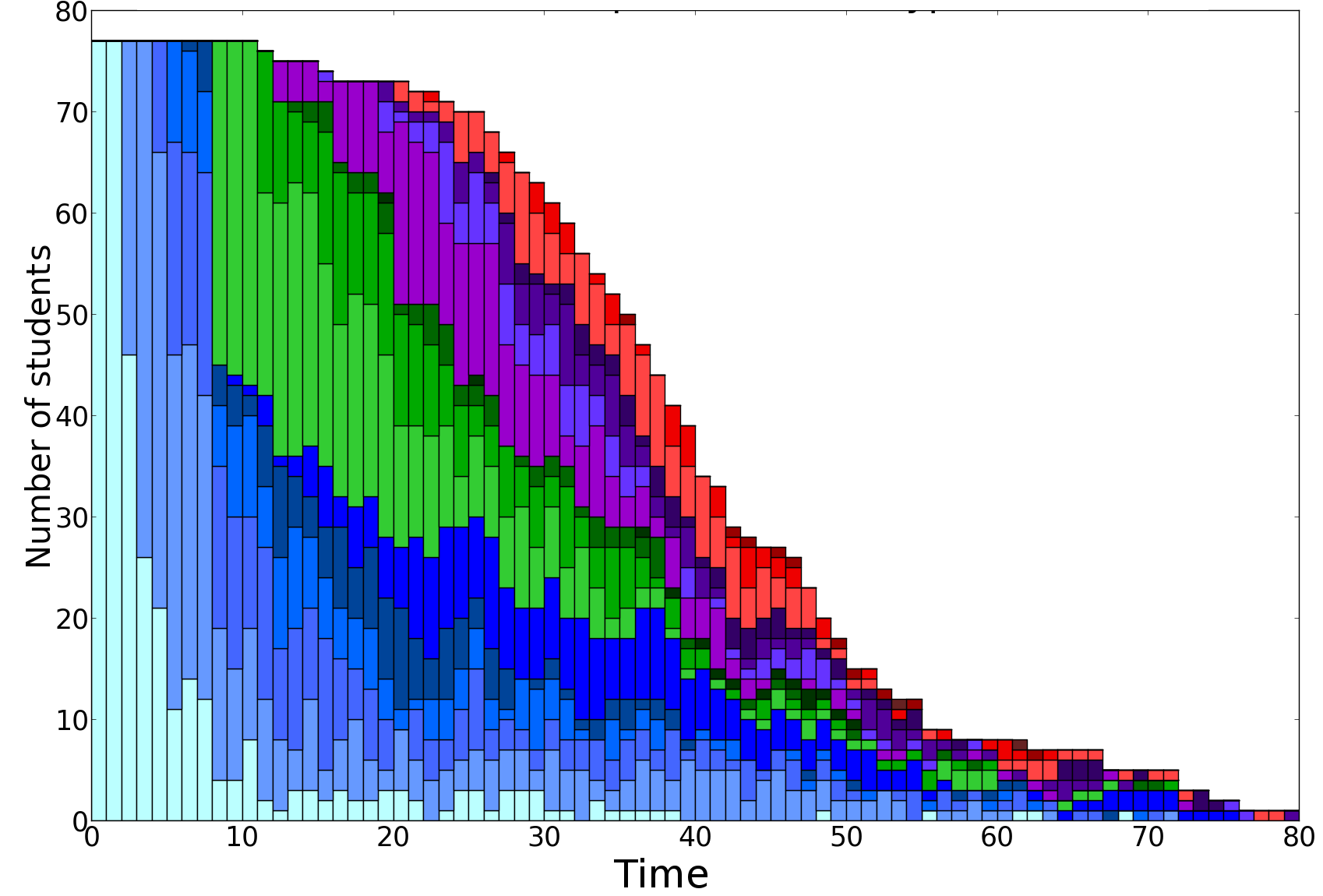}}&
            \includegraphics[width=.40\textwidth]{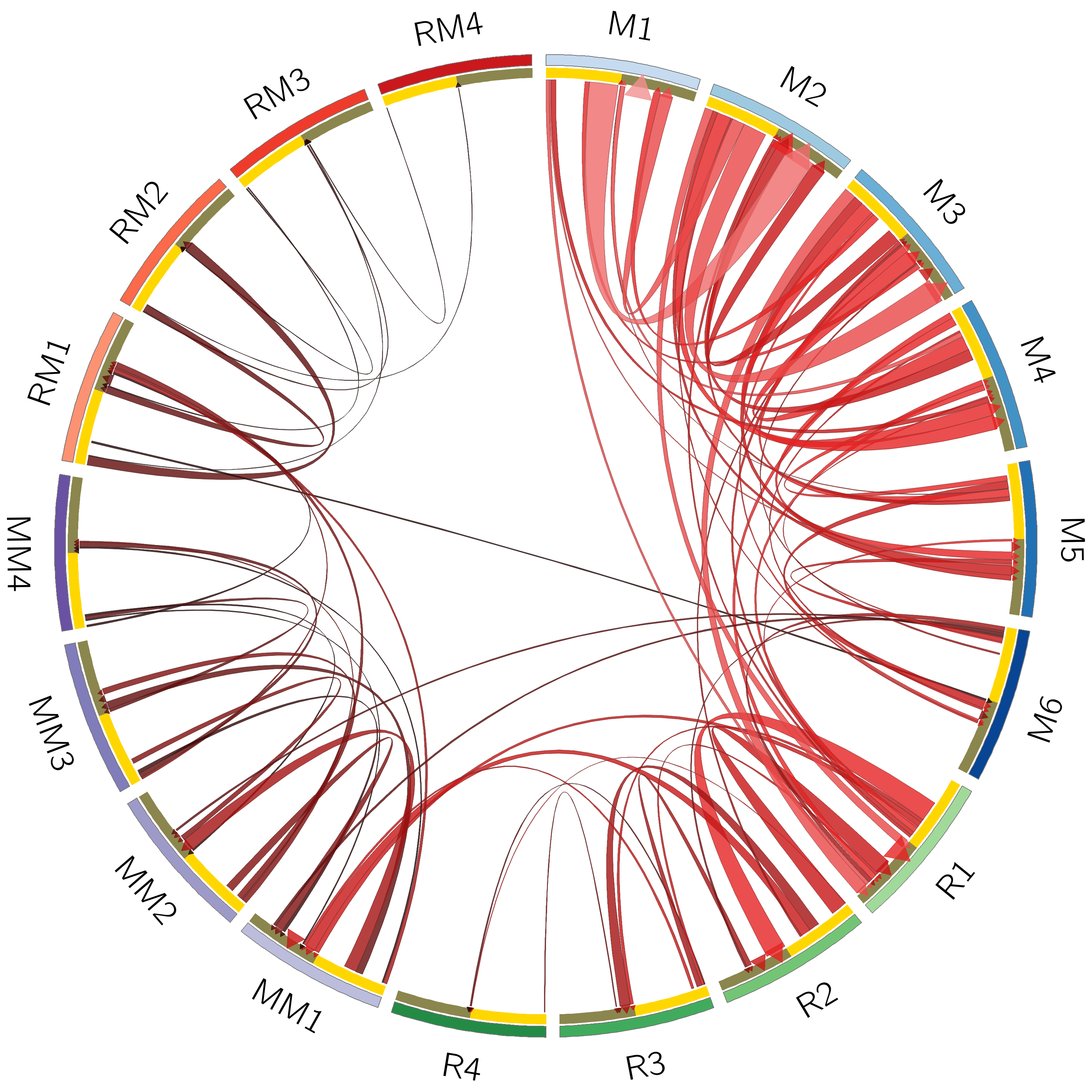}\\
        \end{tabular}
    \caption{For each type and level of exercises : (left) the histograms show the number of students who have achieved it and (right) circos drawing show the number of students doing transition between exercises (pass from one to another), the line thickness represent the number of students who did the transition. Light colors correspond to early exercises, darker colors to later ones. We can see a stronger variety of paths proposed by the automatic algorithms resulting from the online personalization.}
    \label{fig:Levels_timeExp}
\end{figure}

We will now verify if the different algorithms provide qualitatively different learning sequences or if they only adapt the speed of progression. Figure \ref{fig:Levels_timeExp} shows two different things. On the left, the figure shows the evolution of the estimation of the students' competence level, corresponding to the exercise that is being proposed to the learners (only showing the parameter Exercise type and level). On the right side, we can see circular design made using Circos \cite{krzywinski2009circos}. On this figure, the transitions between exercises made by students along time are represented by the colored curved lines (blue for Expert Seq., green for ZPDES and red for RiARiT). A transition starts on an exercise, situated on the yellow part of an exercise, and finish on an other, represented by an arrow and situated on the brown part of an exercise. The line thickness represent the number of students who did that transition. The time is represented by the color shade, light colors correspond to early exercises, darker colors to later ones. This figure allows to see the paths followed by the students for each algorithm. ZPDES even ignored the more difficult exercises of Type M, as it has found that the other types of exercises were providing greater learning progress.

We can see that in general, RiARiT and ZPDES propose a large diversity of type and difficulty of exercises earlier that the Expert Sequence. The same phenomena is visible on Figure~\ref{fig:maxReachAchiev}, where there are more students who reach MM exercises and RM exercises with our algorithms than with the Expert Sequence. The circos figures show that RiARiT and ZPDES proposed more different activities and paths, which reveals an adaptive behavior, where the Expert Sequence proposes almost always the same path.
%

%
%

\subsubsection*{Differences in pre- and post- tests}

\begin{figure}
    \centering
        \begin{tabular}{cc}            
            Control group & Normal group\\
            \includegraphics[width=.49\textwidth]{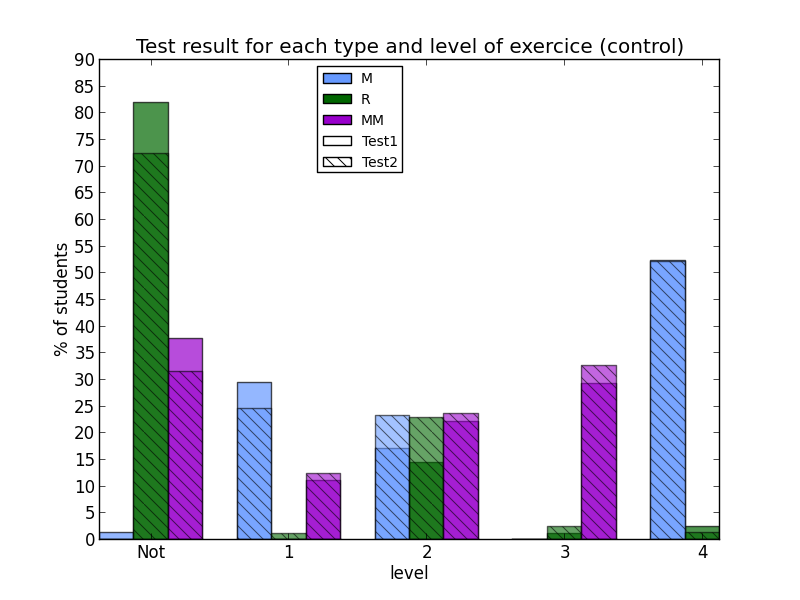}&
            \includegraphics[width=.49\textwidth]{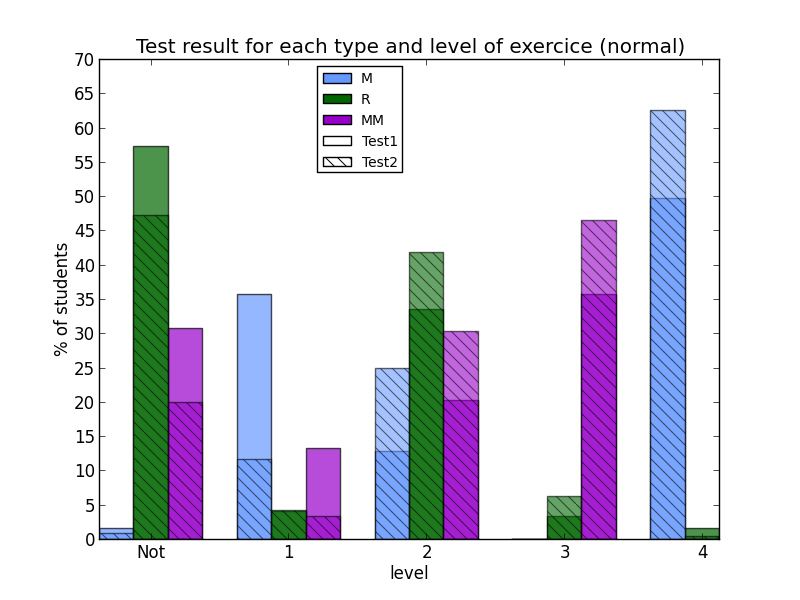}\\  
        \end{tabular}
    \caption{For each type and level of exercises, the histograms show the percentage of students who have achieved it for the first and the second test. For each bar the shaded area corresponds to Test 2. If the shaded goes higher is means that an higher number of students answered correctly in Test 2. Control group is on the left and the normal group is on the right.}
    \label{fig:learnTestControlvsNormal}
\end{figure}

The pre- and post- tests enable to test student knowledge on some KC, buying one object (M) or two (MM) and exercises of giving change (R). To give change with two objects (RM) is not tested because it is not part of the official program for that grade. Figure \ref{fig:learnTestControlvsNormal} shows the evolution of results between pre- and post- tests for the control group (left) which has not used the ITS and the normal group (right). We can see that the normal group improved their results between the pre-test and the post-test, about 65\% of students who were at level 1 for M type are moved to a higher level. Likewise for R and MM types, there are respectively 20\% and 40\% of students who were at level 0 and 1 who have increased their level. For the control group, we can see that their learning is much lower than those who worked on the application. Only 15\% of the students who were at level 0 or 1 for all type of exercises are moved to a higher level. We make an \textit{ANOVA} to test the significance of theses differences. We take as the null hypothesis that the control and the normal group learned the same. We find a $p-value < 5\%$ so we can reject the null hypothesis and we can therefore conclude that the students learned more using the application.

%
%

%
%

\section{Conclusions and Future Work}

In this work, we introduced a new approach for intelligent tutoring systems that relies on multi-armed bandits. Due to their efficiency, these algorithms allow a true personalized learning experience relying on little domain knowledge and doing the optimization online based on the students' results. We introduced a very simple algorithm called ZPDES that relies only on the measure of success and failure on exercises and on a coarse predefined exploration graph to provide a personalized teaching sequence. Another algorithm, RiARiT, is able to exploit more information about the domain to estimate the level of the students and to personalize the teaching sequence. One perhaps not surprising fact is that in simulation RiARiT, with its extra information provided better results, while in the user studies ZPDES provided better results. This result reinforces the conclusion of previous studies that the use of population wide parameters might not be the best thing to do when the goal is to personalize learning.

Our goal is not so much to provide better teaching sequences than expert teachers but, instead, provide a tool that can deliver exercises to the students at their competence level. Nevertheless, our results show that our algorithms can achieve an efficiency of learning comparable to a sequence provided by an expert teacher, even without using much information about the students, and without much information to be provided by the teacher. We showed that we can achieve comparable results for homogeneous populations of students, but a great gain in learning for populations of students  with larger variety and stronger difficulties.

We note that even in cases where there is no gain in learning speed, a formulation of the problem based on the KC is already useful as it identifies more clearly the problems of each particular student, as was observed in the results.

The results from the user studies show a significant increase in learning speed for several competences, and a much better personalization teaching sequence. The algorithm ZPDES is the most promising for a real use as it requires very little information, much less parameters, and has the best adaptation to the user.

~\\

Currently we are studying different teaching scenarios to better identify in which situations our methods provide higher gains and where it can be easily deployed. The advantage of our system is that it has much less assumption in relation to the cognitive and student models, but for this it requires to empirically evaluate the teaching gain of each activity. For this, we expect it to be useful in situations where there are many interactions with the tutoring system and with simpler exercises. It will be more suited to the \textit{inner loop}, i.e. within-activity, of the ITS than to the \textit{outer loop}, i.e. across-activity, see definitions in \citeN{koedinger13ITSsuvery}.
The comparison with other methods is very difficult due to the different assumptions made by each of them. If we have access to a well-identified cognitive/student model for homogeneous populations of students, we might expect approaches based on POMDP to work best. But, for the the more realistic case where the students are more heterogeneous in their levels and learning behaviors, our approach will better address the identifiability problems and the variations in the student population.

Even if our results show that a model might not be the best thing to do when the goal is personalization, the use of accurate methods for learning models, specially the ones using parameters such as \citeN{gonzalez2014general} and \citeN{dhanani2014parameters}, needs still to be better evaluated. A promising approach will consist in using models to bootstrap teaching strategies and then using MABs to personalize to each individual student.

Exploration of further MAB techniques has also to be considered, see \citeN{bubeck2012MAB} for a survey. Possibilities are the use of contextual bandits to take into account the current state of the student and the possible parameters available, and linear bandits to consider more complicated relations between the parameters. A design choice we made was to separate the different parameters into different bandits. This corresponds to consider a factorization on the parameters that is not commonly used. A careful study on the properties of such system will be necessary.

Another interesting direction would be to exploit the clustering that our algorithm implicitly produces in the teaching sequences. We could transfer information from one student to another based on similarities detected at runtime. Methods to exploit transfer in multi-armed-bandits have recently been introduced \cite{Azar13TransferMAB}.

A final point is that we are choosing exercises based on the estimated (recent) past learning progress, and if we know which exercise is next in terms of complexity then we can use that one. This information, if correct, allows the MAB to propose the more complex exercises without requiring to estimate their value first. It also results in a sequence of exercises that is more natural and has less switches between exercises.


\begin{small}
\bibliographystyle{acmtrans}

\end{small}

\newpage
\appendix


\section{Computational Considerations}
\label{sec:ComputationalConsiderations}

In this section we will provide some extra details to explain how we deal with the possible high number of activities. If there are many activities we will need to explore all of them and we will not be able to exploit relations between activities. Also, for a teacher it might be easier to define activities in terms of parameterized activities (or templates as is sometimes called). To address these issue we assume that each activity is represented by a set of $n_p$ parameters $\mathbf{a}=(a^1,...,a^{n_p})$. In this way related activities will have similar parameters. An activity is thus parameterized as follows $\mathbf{a_1}=(Difficulty,Price Presentation,CentNot,RepMoney)$. 
We can no longer consider an activity as a specific combination of parameters because that would leave to a combinatorial explosion. We will thus consider a factorization that instead of using a bandit per activity will use a bandit per parameter value. In the Algorithm~\ref{alg:SSBandit} we need thus to make some small changes. First the $w$ are defined per parameter. As the bandits work at the parameter level, another change is how each exercise is generated. The following lines need to be changed as follows:
\begin{algorithmic}[1]
\setcounter{ALC@line}{8}
\STATE  $p_i = \tilde{w_i} (1-\gamma)+ \gamma \xi_u$
\setcounter{ALC@line}{9}
\STATE Sample $a^i$ proportional to $p_i$ 
\setcounter{ALC@line}{11}
\STATE Propose activity $\mathbf{a}=(a^1,...,a^{n_p})$
\end{algorithmic}

The last change is how the reward is computed. For ZPDES it only means that we will compute the reward and have a specific $w$ per parameter. 

For RiARiT we need to make more changes. The R Table needs also to be factorized. Now each entry on the table is per parameter, where $q_i(a^j)$ denotes the competence level in $KC_j$ required to succeed entirely in activity $\mathbf{a}$ which $j^th$ parameter value is $a^j$, as shown in Table \ref{tab:realQ}. This factorization makes the assumption that activity parameters are not correlated. This assumption is not valid in the general case, but does not change the results in practice. 
We use the factorized R Table in the following manner to heuristically estimate the competence level $q_i(\mathbf{a})$ required in $KC_i$ to succeed in an activity parameterized with $\mathbf{a}$:
\[
    q_i(\mathbf{a}) = \prod_{j=1}^{n_p} q_i(a^j)
\]

\section{Expert Pedagogical Sequence}
\label{sec:PredefinedSeq}

In order to evaluate our algorithm, we use as baseline an optimized sequence created based on instructional design theory, whose reliability has been validated through several user studies, see \cite{roy12math}. This sequence has the following characteristics: 

\begin{itemize}
    \item there is 5 groups for a total of 28 exercises: 
    \begin{itemize}
        \item M exercise with integer price : 3 exercises
        \item M exercise with decimal price : 4 exercises 
        \item R exercise with one object : 5 exercises
        \item MM exercise : 8 exercises
        \item RM exercise : 8 exercises
    \end{itemize} 
    
  \item propose 4 times the same parameterized exercise :
    \begin{itemize}
        \item after 3 correct answers, pass to the next group of exercises. If it's the last exercise group, change the exercise group
        \item else change directly of exercise group  to work on another type of exercise
        \item when changing groups, begin from the highest exercise succeed
    \end{itemize} 
\end{itemize} 

Table~\ref{tab:PredefinedSequence} summarizes the $28$ stages of progression for the students. Following the parameters previously defined, the table also shows how the different parameters evolve. 

\begin{table*}[!htbp]
  \centering
  \caption{Expert sequence including 28 different stages of different parameters for the proposed activities.}
  \small
    \begin{tabular}{|l|c|c|c||c|c|c|c||c|c|c|c|c|}\hline
              & G1.1& G1.2& G1.3& G2.1& G2.2& G2.3& G2.4& G3.1& G3.2& G3.3& G3.4& G3.5  \\\hline
      Ex Type & M   & M   & M   & M   & M   & M   & M   & R   & R   & R   & R   & R     \\\hline
      Difficulty& 1 & 2   & 3   & 4   & 5   & 5   & 6   & 1   & 2   & 3   & 3   & 4     \\\hline
      Cents Not & - & -   & -   & x\euro x & x\euro x & x,x\euro & x,x\euro & - & - & x\euro x &x\euro x& x,x\euro \\\hline
    \end{tabular}

    \vspace{0.5cm}
    
    \begin{tabular}{|l|c|c|c|c|c|c|c|c|}\hline
              & G4.1& G4.2& G4.3& G4.4& G4.5& G4.6& G4.7& G4.8  \\\hline
      Ex Type & MM  & MM  & MM  & MM  & MM  & MM  & MM  & MM    \\\hline
      Difficulty& 1 & 1   & 2   & 2   & 3   & 3   & 4   & 4     \\\hline
      Remainder & - & Int & -   & Int & -   & Int & -   & Dec   \\\hline
      Money Type&Real&Real& Real& Real& Real& Real& Real& Token \\\hline
    \end{tabular}

    \vspace{0.5cm}

    \begin{tabular}{|l|c|c|c|c|c|c|c|c|}\hline
              & G5.1& G5.2& G5.3& G5.4& G5.5& G5.6& G5.7& G5.8  \\\hline
      Ex Type & MM  & MM  & MM  & MM  & MM  & MM  & MM  & MM    \\\hline
      Difficulty& 1 & 1   & 2   & 2   & 3   & 3   & 4   & 4     \\\hline
      Remainder & - & Int & -   & Int & -   & Int & -   & Dec   \\\hline
      Money Type&Real&Real& Real& Real& Real& Real& Real& Token \\\hline
    \end{tabular}
  \label{tab:PredefinedSequence}
\end{table*}

\section{Tables}

The following tables (\ref{tab:realQ}, \ref{tab:pereq}, \ref{tab:desactivation}) provide all the parameters during the user studies when using the algorithm RiARiT.

\begin{table*}[htbp]
    \centering
    \caption{R Tables that was used in the user study for algorithm RiARiT. It shows the relation of the parameters values and the minimum required competence level, for each KC, to be able to solve that exercise.}
    \label{tab:realQ}
        \begin{tabular}{|l|c|c|c|c|c|c|c|c|c|c|c|}\hline
                &&KnowMoney&IntSum&IntSub&IntDec&DecSum&DecSub&DecDec\\\hline
\multirow{4}{*}{Exercise Type}  & M   &  1  &  0.5  &  0.3  &  0.7  &  0.3  &  0.2  &  0.7  \\
                                & R   &  1  &  0.5  &  0.8  &  0.7  &  0.3  &  0.7  &  0.7  \\
                                & MM  &  1  &  1    &  0.4  &  1    &  1    &  0.3  &  1    \\
                                & RM  &  1  &  1    &  1    &  1    &  1    &  1    &  1    \\\hline
        \end{tabular}

        \vspace{0.5cm}

        \begin{tabular}{|l|c|c|c|c|c|c|c|c|c|c|c|}\hline
                &&KnowMoney&IntSum&IntSub&IntDec&DecSum&DecSub&DecDec\\\hline
\multirow{6}{*}{M difficulty}   &  1  &  0.3  &  0.2  &  0    &  0    &  0    &  0    &  0   \\
                                &  2  &  0.5  &  0.5  &  0.3  &  0.5  &  0    &  0    &  0   \\
                                &  3  &  0.5  &  0.6  &  0.5  &  0.7  &  0    &  0    &  0   \\
                                &  4  &  0.7  &  0.4  &  0    &  0    &  0    &  0    &  0.3 \\
                                &  5  &  0.9  &  0.8  &  0.7  &  0.7  &  0.5  &  0.6  &  0.6 \\
                                &  6  &  1    &  1    &  1    &  1    &  1    &  1    &  1   \\\hline
        \end{tabular}

        \vspace{0.5cm}

        \begin{tabular}{|l|c|c|c|c|c|c|c|c|c|c|c|}\hline
                &&KnowMoney&IntSum&IntSub&IntDec&DecSum&DecSub&DecDec\\\hline
\multirow{4}{*}{R difficulty}   &  1  &  0.3  &  0.6  &  0.4  &  0.6  &  0    & 0    &  0    \\
                                &  2  &  0.5  &  1    &  0.7  &  1    &  0    & 0    &  0    \\
                                &  3  &  0.8  &  0.8  &  0.9  &  0.8  &  0.5  & 0.5  &  0.5  \\
                                &  4  &  1    &  1    &  1    &  1    &  1    & 1    &  1    \\\hline
        \end{tabular}

        \vspace{0.5cm}

        \begin{tabular}{|l|c|c|c|c|c|c|c|c|c|c|c|}\hline
                &&KnowMoney&IntSum&IntSub&IntDec&DecSum&DecSub&DecDec\\\hline
\multirow{4}{*}{MM difficulty}  &  1  &  0.5  &  0.6  &  1  &  1  &  0    &  0  &  0    \\
                                &  2  &  0.5  &  0.7  &  1  &  1  &  0    &  0  &  0    \\
                                &  3  &  0.8  &  1    &  1  &  1    &  0.7  &  1  &  0.8  \\
                                &  4  &  1    &  1    &  1  &  1    &  1    &  1  &  1    \\\hline
        \end{tabular}

        \vspace{0.5cm}

        \begin{tabular}{|l|c|c|c|c|c|c|c|c|c|c|c|}\hline
                &&KnowMoney&IntSum&IntSub&IntDec&DecSum&DecSub&DecDec\\\hline
\multirow{4}{*}{RM difficulty}  &  1  &  0.5  &  0.6  &  0.7  &  1  &  0    &  0    &  0    \\
                                &  2  &  0.5  &  0.7  &  0.7  &  1  &  0    &  0    &  0    \\
                                &  3  &  0.8  &  1    &  0.8  &  1  &  0.7  &  0.7  &  0.7  \\
                                &  4  &  1    &  1    &  1    &  1  &  1    &  1    &  1    \\\hline
        \end{tabular}

        \vspace{0.5cm}

        \begin{tabular}{|l|c|c|c|c|c|c|c|c|c|c|c|}\hline
                &&KnowMoney&IntSum&IntSub&IntDec&DecSum&DecSub&DecDec\\\hline
\multirow{2}{*}{M/R modality}   & x\euro x  &  0.8  &  1   &  1   &  1  &  0.6  &  1    &  0.7    \\
                                & x.x\euro  &  1    &  1   &  1   &  1  &  1    &  1    &  1    \\\hline
        \end{tabular}

        \vspace{0.5cm}

        \begin{tabular}{|l|c|c|c|c|c|c|c|c|c|c|c|}\hline
                &&KnowMoney&IntSum&IntSub&IntDec&DecSum&DecSub&DecDec\\\hline
\multirow{3}{*}{Remainder}      &  No       &  1    &  0.7  &  1    &  1  &  0.7  &  1    &  1    \\
                                &  Unit     &  1    &  1    &  1    &  1  &  0.7  &  1    &  1    \\
                                &  Decimal  &  1    &  1    &  1    &  1  &  1    &  1    &  1    \\\hline

\multirow{2}{*}{Money Type}     &  Real     &  1    &  1  &  1  &  1  &  1    &  1    &  0.8    \\
                                &  Token    &  0.9  &  1  &  1  &  1  &  1    &  1    &  1    \\\hline
        \end{tabular}

\end{table*}

\begin{table*}[htbp]
    \centering
    \caption{This table shows the pedagogical restrictions that are enforced when using the RiARiT algorithm. A given exercise parameter can only be used if the pre-condition is achieved, usually in the form of a minimum skill level for a given KC.}
    \label{tab:pereq}
        \begin{tabular}{|l|c|c|c|c|c|c|c|c|c|c|c|}\hline
                &&KnowMoney&IntSum&IntSub&IntDec&DecSum&DecSub&DecDec\\\hline
\multirow{4}{*}{Exercise Type}  & M   &  0    &  0    &  0    &  0    &  0    &  0    &  0  \\
                                & R   &  0.3  &  0.2  &  0    &  0.3  &  0  &  0  &  0  \\
                                & MM  &  0.3  &  0.3  &  0.3  &  0.3  &  0  &  0  &  0  \\
                                & RM  &  0.3  &  0.5  &  0.3  &  0.3  &  0  &  0  &  0  \\\hline
        \end{tabular}

        \vspace{0.5cm}

        \begin{tabular}{|l|c|c|c|c|c|c|c|c|c|c|c|}\hline
                &&KnowMoney&IntSum&IntSub&IntDec&DecSum&DecSub&DecDec\\\hline
\multirow{6}{*}{M difficulty}   &  1  &  0    &  0    &  0    &  0    &  0  &  0  &  0    \\
                                &  2  &  0.1  &  0.1  &  0    &  0    &  0  &  0  &  0    \\
                                &  3  &  0    &  0    &  0    &  0.3  &  0  &  0  &  0    \\
                                &  4  &  0.3  &  0.3  &  0.2  &  0    &  0  &  0  &  0    \\
                                &  5  &  0    &  0    &  0    &  0    &  0  &  0  &  0.1  \\
                                &  6  &  0    &  0    &  0    &  0    &  0  &  0  &  0.4  \\\hline
        \end{tabular}

        \vspace{0.5cm}

        \begin{tabular}{|l|c|c|c|c|c|c|c|c|c|c|c|}\hline
                &&KnowMoney&IntSum&IntSub&IntDec&DecSum&DecSub&DecDec\\\hline
\multirow{4}{*}{R difficulty}   &  1  &  0    &  0  &  0    &  0  &  0  &  0    &  0    \\
                                &  2  &  0    &  0  &  0.3  &  0  &  0  &  0    &  0    \\
                                &  3  &  0.4  &  0  &  0.5  &  0  &  0  &  0    &  0    \\
                                &  4  &  0.6  &  0  &  0.6  &  0  &  0  &  0.3  &  0.3  \\\hline
        \end{tabular}

        \vspace{0.5cm}

        \begin{tabular}{|l|c|c|c|c|c|c|c|c|c|c|c|}\hline
                &&KnowMoney&IntSum&IntSub&IntDec&DecSum&DecSub&DecDec\\\hline
\multirow{4}{*}{MM difficulty}  &  1  &  0  &  0    &  0  &  0  &  0  &  0  &  0  \\
                                &  2  &  0  &  0.3  &  0  &  0  &  0  &  0  &  0  \\
                                &  3  &  0  &  0.4  &  0  &  0  &  0  &  0  &  0    \\
                                &  4  &  0  &  0    &  0  &  0  &  0  &  0  &  0.5  \\\hline
        \end{tabular}

        \vspace{0.5cm}

        \begin{tabular}{|l|c|c|c|c|c|c|c|c|c|c|c|}\hline
                &&KnowMoney&IntSum&IntSub&IntDec&DecSum&DecSub&DecDec\\\hline
\multirow{4}{*}{RM difficulty}  &  1  &  0  &  0    &  0    &  0  &  0  &  0    &  0  \\
                                &  2  &  0  &  0    &  0.4  &  0  &  0  &  0    &  0  \\
                                &  3  &  0  &  0.6  &  0    &  0  &  0  &  0    &  0  \\
                                &  4  &  0  &  0.7  &  0    &  0  &  0  &  0.4  &  0  \\\hline
        \end{tabular}

        \vspace{0.5cm}

        \begin{tabular}{|l|c|c|c|c|c|c|c|c|c|c|c|}\hline
                &&KnowMoney&IntSum&IntSub&IntDec&DecSum&DecSub&DecDec\\\hline
\multirow{3}{*}{Remainder}      &  No       &  0    &  0    &  0    &  0  &  0    &  0    &  0    \\
                                &  Unit     &  0    &  0.4  &  0    &  0  &  0    &  0    &  0    \\
                                &  Decimal  &  0    &  0    &  0    &  0  &  0    &  0    &  0    \\\hline
        \end{tabular}
\end{table*}

\begin{table*}[htbp]
    \centering
    \caption{This table shows another type of pedagogical restrictions that are enforced into the RiARiT algorithm. A given exercise parameter can be deactivated if the pre-condition is achieved, usually in the form of maximum skill levels for one or many KC.}
    \label{tab:desactivation}

        \begin{tabular}{|l|c|c|c|c|c|c|c|c|c|c|c|}\hline
                &&KnowMoney&IntSum&IntSub&IntDec&DecSum&DecSub&DecDec\\\hline
\multirow{6}{*}{M difficulty}   &  1  &  0  &  0.6  &  0  &  0    &  0  &  0  &  0    \\
                                &  2  &  0  &  0    &  0  &  0.7  &  0  &  0  &  0    \\
                                &  3  &  0  &  0    &  0  &  0    &  0  &  0  &  0.8  \\
                                &  4  &  0  &  0    &  0  &  0    &  0  &  0  &  0.7  \\
                                &  5  &  0  &  0    &  0  &  0    &  0  &  0  &  0.8  \\
                                &  6  &  1  &  1    &  1  &  1    &  1  &  1  &  1    \\\hline
        \end{tabular}

        \vspace{0.5cm}

        \begin{tabular}{|l|c|c|c|c|c|c|c|c|c|c|c|}\hline
                &&KnowMoney&IntSum&IntSub&IntDec&DecSum&DecSub&DecDec\\\hline
\multirow{4}{*}{R difficulty}   &  1  &  0  &  0  &  0  &  0.7  &  0    &  0  &  0   \\
                                &  2  &  0  &  0  &  0  &  0    &  0.7  &  0  &  0   \\
                                &  3  &  0  &  0  &  0  &  0    &  0    &  0  &  0.8 \\
                                &  4  &  1  &  1  &  1  &  1    &  1    &  1  &  1   \\\hline
        \end{tabular}

        \vspace{0.5cm}

        \begin{tabular}{|l|c|c|c|c|c|c|c|c|c|c|c|}\hline
                &&KnowMoney&IntSum&IntSub&IntDec&DecSum&DecSub&DecDec\\\hline
\multirow{4}{*}{MM difficulty}  &  1  &  0  &  0.8  &  0  &  0  &  0  &  0    &  0  \\
                                &  2  &  0  &  0.9  &  0  &  0  &  0  &  0    &  0  \\
                                &  3  &  0  &  0    &  0  &  0  &  0  &  0.9  &  0  \\
                                &  4  &  1  &  1    &  1  &  1  &  1  &  1    &  1  \\\hline
        \end{tabular}

        \vspace{0.5cm}

        \begin{tabular}{|l|c|c|c|c|c|c|c|c|c|c|c|}\hline
                &&KnowMoney&IntSum&IntSub&IntDec&DecSum&DecSub&DecDec\\\hline
\multirow{4}{*}{RM difficulty}  &  1  &  0  &  0  &  0.8  &  0  &  0  &  0    & 0  \\
                                &  2  &  0  &  0  &  0    &  0  &  0  &  0.8  & 0  \\
                                &  3  &  0  &  0  &  0    &  0  &  0  &  0.9  & 0  \\
                                &  4  &  1  &  1  &  1    &  1  &  1  &  1    & 1  \\\hline
        \end{tabular}
\end{table*}

\end{document}